\documentclass[accepted]{uai2021} % after acceptance, for a revised
\usepackage[american]{babel}
\usepackage{natbib} % has a nice set of citation styles and commands
\usepackage{mathtools} % amsmath with fixes and additions
\usepackage{booktabs} % commands to create good-looking tables
\usepackage{tikz} % nice language for creating drawings and diagrams

\usepackage{xcolor}
\usepackage{amsmath,amssymb,mathrsfs}
\usepackage{amsthm}
\usepackage{epsfig,graphics}
\usepackage{color}

\usepackage{algpseudocode}
\usepackage{algorithm}
\usepackage{microtype}
\usepackage{graphicx}
\usepackage{enumitem}

\newcommand{\red}[1]{\textcolor{red}{#1}}
\definecolor{darkspringgreen}{rgb}{0.09, 0.45, 0.27}
\newcommand{\green}[1]{\textcolor{darkspringgreen}{#1}}

\newcommand{\orange}[1]{\textcolor{orange}{#1}}
\newcommand{\blue}[1]{\textcolor{blue}{#1}}

\def\rd{{\rm d}}

\def\argmin{{\rm argmin}}

\def\KL{{\rm KL}}

\def\E{{\mathbb{E}}}
\DeclareMathOperator\supp{supp}
\usepackage{pifont}
\newcommand{\cmark}{\text{{\small \ding{51}}}}
\newcommand{\xmark}{\text{{\small \ding{55}}}}

\usepackage{xr}

\makeatletter
\newcommand*{\addFileDependency}[1]{% argument=file name and extension
  \typeout{(#1)}
  \@addtofilelist{#1}
  \IfFileExists{#1}{}{\typeout{No file #1.}}
}
\makeatother

\title{Variational Refinement for Importance Sampling \\ Using the Forward Kullback-Leibler Divergence}

\author[*1,2,3]{\href{mailto:Ghassen Jerfel <ghassen@google.com>?Subject=UAI 2021 Variational Refinement for Importance Sampling Using the Forward Kullback-Leibler Divergence}{Ghassen Jerfel}{}}
\author[*1,2]{\href{mailto:Serena Wang <serenawang@google.com>?Subject=UAI 2021 Variational Refinement for Importance Sampling Using the Forward Kullback-Leibler Divergence}{Serena Wang}{}}
\author[2]{Clara Fannjiang}
\author[1,3]{Katherine A. Heller}
\author[1,2,4]{Yian Ma}
\author[2]{\\Michael I. Jordan}
\affil[1]{%
    Google Research\\ 
    Mountain View, CA, USA
}
\affil[2]{%
    University of California Berkeley\\
    Berkeley, CA, USA
}
\affil[3]{%
    Duke University\\
    Durham, NC, USA
}
\affil[4]{%
    University of California San Diego\\
    San Diego, CA, USA
}

\begin{document}

\maketitle

\begin{abstract}
Variational Inference (VI) is a popular alternative to asymptotically exact sampling in Bayesian inference. Its main workhorse is optimization over a reverse Kullback-Leibler divergence (RKL), which typically underestimates the tail of the posterior leading to miscalibration and potential degeneracy. 
Importance sampling (IS), on the other hand, is often used to fine-tune and de-bias the estimates of approximate Bayesian inference procedures. 
The quality of IS crucially depends on the choice of the proposal distribution. 
% Coincidentally, underestimation of the tail in the proposal is also a serious hindrance of its application.
Ideally, the proposal distribution has heavier tails than the target, which is rarely achievable by minimizing the RKL.
We thus propose a novel combination of optimization and sampling techniques for approximate Bayesian inference by constructing an IS proposal distribution through the minimization of a forward KL (FKL) divergence. 
This approach guarantees asymptotic consistency and a fast convergence towards both the optimal IS estimator and the optimal variational approximation.
We empirically demonstrate on real data that our method is competitive with variational boosting and MCMC.
\end{abstract}

\section{Introduction}
Bayesian analysis provides a principled framework to encode complex hierarchical structures and prior beliefs in order to capture posterior uncertainty about latent variables $\theta$ given observed data $x$ via the posterior $p(\theta|x)$. The inferential goal often involves computing expectations over this posterior distribution, $\E_{\theta \sim p(\theta|x)}[f(\theta)]$, which is typically accomplished by sampling.
%
% \red{Bayesian analysis provides a powerful framework to encode complex hierarchical structures and prior beliefs in order to coherently capture posterior uncertainty ($p(\theta|x)$) about the latent variables $\theta$ given the observed variables $x$.} Posterior samples are then typically used to construct statistical estimators from summaries of interest such as expectations of desired quantities, credible intervals, and probabilities of rare events. 
%
Unfortunately, sampling directly from the posterior is usually intractable. Computing posterior functionals thus requires approximate inference methods such as variational inference (VI) \citep{jordan1999,wainwright2008}, Markov Chain Monte Carlo (MCMC) \citep{brooks2011,andrieu2003}, and importance sampling (IS) \citep{gelman1998simulating}, among others.
%which can be estimated by sampling from the posterior.
%In most applications,  the posterior distribution is difficult compute due to an intractable marginal likelihood $p(x)$. 
% This has spurred the rise of approximate Bayesian inference methods such as variational inference (VI) \citep{blei2016,jordan1999,wainwright2008}, and Markov Chain Monte Carlo (MCMC) \red{[ref]} methods which include importance sampling (IS) \red{[ref]}.

Recently, variational inference has grown in popularity because it recasts the inference problem as an optimization problem that can leverage recent advances in stochastic optimization \citep{bottou2010,hoffman2013} and automatic differentiation \citep{maclaurin2015}.
Specifically, VI poses a tractable family of distributions $\mathcal{Q}$ and minimizes the reverse Kullback-Leibler divergence (RKL) between $q$ and $p$, i.e. $\KL(q||p)$.
However, $\mathcal{Q}$ is often misspecified leading to unknown bias in VI solutions~\citep{blei2016}. It is also generally difficult to assess the quality of a VI approximation on downstream tasks based on the value of the RKL divergence \citep{yao2018yes,campbell2019universal,rainforth2018tighter,huggins2020validated}.
This motivates the use of importance sampling (IS) to de-bias posterior summaries regardless of the misspecification of $\mathcal{Q}$~\citep{gelman1998simulating,owen2013monte}. 
%With IS, one can de-bias posterior summaries from a variational approximation $q(\theta)$ at the cost of added variance relative to a simple Monte Carlo estimate. 
Unlike the RKL,
the performance of $q$ as an IS proposal is indicative of its quality as an approximation of $p$ on downstream tasks~\citep{yao2018yes}.

This workflow capitalizes on complementary strengths: the computational efficiency of VI can sidestep the challenge of selecting a proposal distribution for IS which, in turn, ensures the consistency of the refined posterior expectation. However, RKL minimization typically results in light tails which can cause instability for importance sampling \citep{dieng2017variational, yao2018yes}. On the other hand, the forward KL divergence (FKL or $\KL(p||q)$) is known to control the estimation error of importance sampling \citep{chatterjee2018sample} but is rarely used due to its intractability. 
%Furthermore, forward KL does not suffer from issues of covariance underestimation \citep{murphy2012machine,campbell2019universal} or zero forcing \citep{hoffman2017learning}.

In this paper, we propose to replace reverse KL with forward KL as the variational objective whose minimization yields an optimal IS proposal distribution. % even in settings where both IS and VI are known to struggle (Table \ref{tab:comp})
We make four distinct contributions in this vein:
\begin{enumerate}[nolistsep]
\item We derive a self-normalized importance sampling estimate for the intractable FKL divergence. 
\item We demonstrate how FKL-based boosting can combine IS and VI for multimodal target distributions. 
\item We show that FKL boosting is guaranteed to converge at a rate of $O(\frac{1}{K})$, where K is the number of boosting iterations, to the best approximation from a family of mixture distributions. This immediately guarantees convergence to the optimal proposal distribution as per the results of \citep{chatterjee2018sample}.
\item We demonstrate empirically that our approach is competitive with state-of-the-art VI and Hamiltonian Monte Carlo on regression tasks over real datasets using Bayesian neural networks (BNNs) and Bayesian linear regression (BLR). 
\end{enumerate}
Our proposed algorithm is thus a principled inference technique, with a well-defined computation-quality trade-off, that can be used independently or as a refining step to correct for the error in a given approximation.
\begin{figure}[!ht]
\centering
\begin{tabular}{c}
Minimize Reverse KL: $\KL(q\|p) = \E_{q} [\log{\frac{q}{p}}]$ \\
\includegraphics[width=0.8\linewidth]{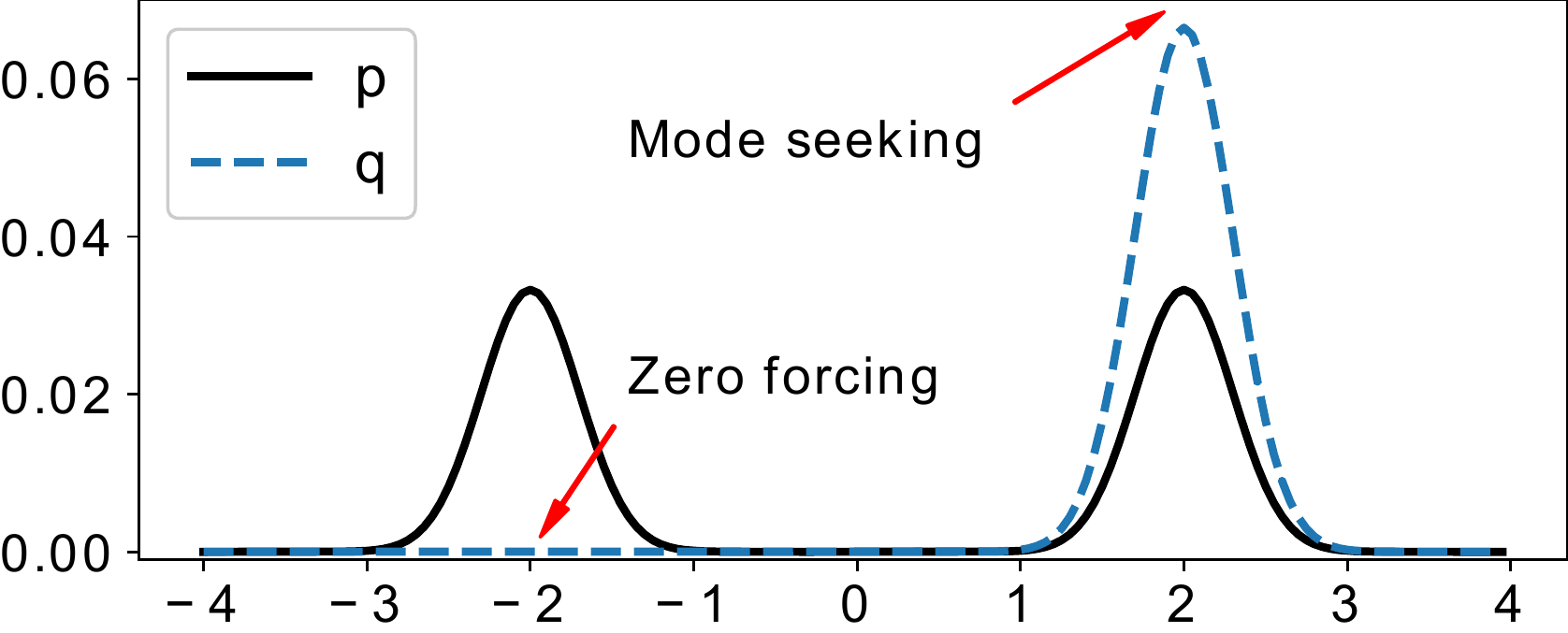} \\
Minimize Forward KL: $\KL(p\|q) = \E_{p} [\log{\frac{p}{q}}]$ \\
\includegraphics[width=0.8\linewidth]{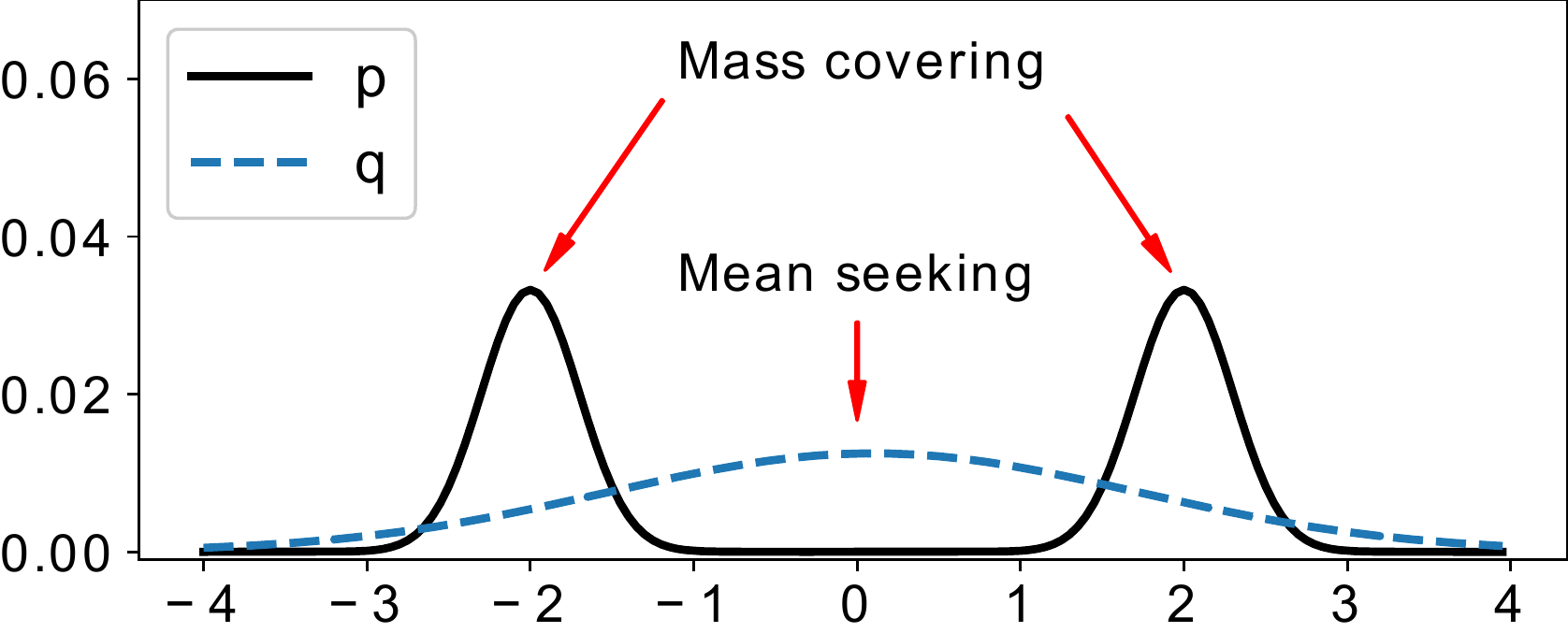} \\
\end{tabular}
\caption{Results of minimizing RKL vs.\ FKL for a Gaussian variational approximation on a bimodal target.}
\label{fig:kl_comp}
\end{figure}
\section{Background}
\label{sec:background}
Let $\theta$ denote the variable of interest with probability density $p$, and let $f$ denote a function of $\theta$. Our goal is to estimate an expectation
\begin{equation}
\E_{\theta \sim p(\theta|x)} \left[ f(\theta) \right].
\label{eq:expectation}
\end{equation}
%where we refer to $p$ as the target distribution. 
In Bayesian inference, $\theta$ generally represents a latent variable to be integrated over in the posterior distribution, $p(\theta | x)$, conditioned on the observed data $x$.

If one can draw $S$ samples $\{\theta_s\}_{s=1}^S$ from the target distribution $p(\theta|x)$ then Eq.~\eqref{eq:expectation} can be estimated by simple Monte Carlo integration. % TODO: should we include the expectation here ?
However, $p(\theta|x)$ is typically only known up to a normalization constant and thus cannot be readily sampled from.
Instead, given samples from an approximation $q$ of the posterior, we can estimate the expectation as:
\begin{equation}
\E_{p(\theta|x)} [f(\theta)] \approx \frac{\sum_{i=1}^S f(\theta_s) w_s}{\sum_{i=1}^S w_s}.
\label{eq:mc_expectation}
\end{equation}
If the samples are weighted equally (i.e., $w_s=1$), Eq.~\eqref{eq:mc_expectation} is equivalent to the VI estimate, which has low variance but can be biased and inconsistent~\citep{owen2013monte}.

If instead we weigh by the importance ratios, $w_s = \frac{p(\theta_s|x)}{q(\theta_s)}$, we recover the IS estimate:
\begin{equation}
\E_{\theta \sim p(\theta|x)} \left[ f(\theta) \right] = \E_{\theta \sim q(\theta)} \left[ \frac{p(\theta|x)}{q(\theta)}f(\theta) \right],
\label{eq:importance_sampling}
\end{equation}
which is consistent (bias=$\mathcal{O}(1/S)$) but with potentially large or infinite variance~\citep{owen2013monte}.
\subsection{VI with Reverse KL Minimization}
Variational inference posits a family $\mathcal{Q}$ of distributions that are easy to evaluate or sample from, and defines the variational approximation $q^* \in \mathcal{Q}$ as the distribution that minimizes the reverse Kullback-Leibler (RKL) divergence to the posterior $p$: $q^* = \underset{q \in \mathcal{Q}}{\argmin} \quad \KL(q\|p)$.

Unless the choice of $\mathcal{Q}$ specifically includes the target distribution, minimizing the reverse KL divergences leads to a $q^*$ that underestimates the target covariance. The decomposition of this divergence sheds light on the cause:
% The decomposition of this divergence sheds light on the cause of covariance underestimation when minimizing the RKL, regardless of the choice of family $\mathcal{Q}$: 
\begin{equation}
    \KL(q\|p) = \E_q [\log{q}] - \E_q [\log{p}]
\label{eq:rkl}
\end{equation} The first term in Eq.~\eqref{eq:rkl} is the entropy of $q$ whose penalization is known to cause light tails. Furthermore, the second term is minimized when $q = 0$ for $p > 0$, leading to \textit{zero forcing} or over-pruning~\citep{higgins2016beta}.
The RKL divergence thus favors a single mode (\textit{mode seeking}), is biased towards avoiding false positives, and poses difficulties when approximating heavy-tailed \citep{guo2016boosting,li2016renyi,dieng2017variational} or multimodal targets \citep{miller2017}. This effect is illustrated in Figure \ref{fig:kl_comp}.
\subsection{Importance Sampling and FKL}
Due to the unknown bias of an RKL-minimizing $q^*$, refining through importance weighting is recommended to de-bias the estimate of Eq.~\eqref{eq:expectation} \citep{yao2018yes,vehtari2015pareto}. However, the light tails of this $q^*$ can lead to high or even infinite variance for an IS estimator which limits the efficiency of $q^*$ as an IS proposal distribution.

As an alternative to RKL, minimizing the forward KL divergence can mitigate the issue of covariance underestimation and tail undersampling. Consider the decomposition of the forward KL divergence: 
\begin{equation}
\KL(p\|q) = \E_p [\log{p}] - \E_p [\log{q}].
\label{eq:fkl_decomp}
\end{equation}
In Eq.~\eqref{eq:fkl_decomp}, the cross-entropy is minimized by setting $q>0$ whenever $p>0$, which leads to \textit{mass covering} as illustrated in Figure \ref{fig:kl_comp}. With better tail coverage, an FKL-minimizing $q^*$ can yield an IS estimate (Eq.~ \eqref{eq:importance_sampling}) with lower variance than an approximation generated by RKL. Indeed, ~\cite{chatterjee2018sample} demonstrate that the variance of an importance sampling estimate scales as $O\left( \frac{e^{\KL(p\|q)}}{\sqrt{S}} \right),$
where the number of samples $S$ required for IS to provide accurate mean estimates scales exponentially with FKL.

\section{Related work}
\label{sec:related}
We present related work on improving estimates of the expectation over intractable target distributions (e.g. high dimensional, heavy tailed, or multimodal).
\subsection{Variational Inference Divergences} 
Prior work has addressed the covariance underestimation and light tails pathologies of reverse KL minimization while seeking to improve the quality of the approximation as an IS proposal through the minimization of alternative divergences such as (reversed) Renyi-$\alpha$ \citep{li2016renyi, hernandez-gonzales2020}, Chi-square ($\alpha=2$) \citep{dieng2017variational}, or Hellinger ($\alpha=1/2$) \citep{campbell2019universal} divergences. FKL can be seen as a special case of $\alpha$ divergences when $\alpha \to 1$ which is not considered in any of these prior VI works.
\subsection{FKL for Approximate Inference} 
While the forward KL's computational inconvenience has limited its use for variational inference, inference techniques such as Belief Propagation (BP) and Expectation Propagation (EP) can be regarded as performing FKL minimization locally~\citep{minka2005divergence}: KL is minimized one data partition at a time instead of globally as in VI. \cite{minka2001family} demonstrates that this local minimization procedure is not guaranteed to converge and may not result in representative posteriors.
%not guaranteed to converge. Furthermore, EP/BP does not provide an easy estimate or bound on the marginals.
%
Another set of techniques that utilize variants of the FKL divergence includes reweighted wake-sleep~\citep{bornschein2014reweighted} which alternates minimizing an approximation of FKL during the sleep phase while minimizing an approximation of RKL during the wake phase. However, this is known to lead to a biased estimator~\citep{bornschein2014reweighted}.
% \red{TODO: more refs about re-weighted wake-sleep / neural sequential monte carlo}
\subsection{Variational Boosting}
Variational boosting (VB)~\citep{miller2017,guo2016boosting,locatello2018boosting, jerfel2017boosted} has been suggested in various forms to address the multimodality challenge for variational inference.  Variational boosting posits a family of mixture distributions $\mathcal{Q}_k$:
\begin{equation}
\mathcal{Q}_k = \left\{q: q(\theta) = \sum_{i=1}^k \lambda_i f_i(\theta), {\lambda}\in\Delta_k \right\},
\label{eq:mixture_family}
\end{equation}
% {\footnotesize
% \begin{equation}
% \mathcal{Q}_k = \left\{q: q(\theta) = \sum_{i=1}^k \lambda_i f_{\phi_j}(\theta), {\lambda}\in\Delta_k, \phi\in\Phi^k \right\},
% \label{eq:mixture_family}
% \end{equation}
% }
and sequentially constructs a variational mixture approximation by adding and re-weighting one (typically Gaussian) mixture component at a time to minimize the KL objective: 
\begin{align}
    &\{\mu_i\}_{i=1}^k, \{\Sigma_i\}_{i=1}^k, \{\lambda_i\}_{i=1}^k \nonumber \\
    &\quad\gets \underset{\mu, \Sigma, \lambda}{\argmin} \; \KL\left( \sum_{i=1}^k \lambda_i f_i(\theta; \mu_i, \Sigma_i) \bigg\| p(\theta|x)\right).
\end{align}
This form of reverse KL-based boosting is known to struggle with degeneracy where the optimization at certain boosting iterations can lead to point-mass mixture components \citep{campbell2019universal}. Ad-hoc regularization techniques are often needed in practice \citep{locatello2018boosting}, but are not necessarily sufficient \citep{campbell2019universal}.
% Variational boosting can be regarded as a gradient boosting technique that attempts to minimize a function residual at each iteration \cite{guo2016boosting,jerfel2017boosted}. For RKL-based boosting, minimizing the residual is equivalent to minimizing the inner product $\langle f_i, \log{(q_{i-1}/p)} \rangle$ which is ill-posed since $f_i$ will degenerate to a point mass at the mode of the residual $\log{(q_{i-1}/p)}$.
To address this pathology, \citet{campbell2019universal} proposed a boosting algorithm, based on the Hellinger divergence, which does not not guarantee scalability with dimensions.
%unlike with reverse KL boosting. % this is risky/unverified claim
\subsection{Adaptive Importance Sampling}
Adaptive IS (AIS) methods such as Adaptive multiple IS \citep{cornuet2012adaptive} and incremental mixture IS \citep{raftery2010estimating} are designed for multimodal targets. However, none of the existing works directly optimize for the FKL divergence which controls the worst case IS estimation error. For example, \citet{cappe2008adaptive} minimize an entropy criterion whereas \citet{douc2007minimum} minimize the empirical variance of the importance weights, which does not necessarily correlate with the quality of IS estimation \citep{vehtari2015pareto}. 

\subsection{Combining IS and VI}
As outlined in Table \ref{tab:comp}, VI and VB suffer from covariance underestimation, and can struggle to approximate heavy-tailed distributions~\citep{blei2016}. AIS, on the other hand, can approximate heavy-tailed targets but cannot scale efficiently in dimensionality~\citep{owen2013monte}. A combination of these two lines of research may benefit from their complementary strengths while sidestepping shared weaknesses (e.g., multimodality). However, it is often difficult to combine optimization-based and sampling-based inference techniques. This is because sampling methods such as MCMC define the approximate distribution implicitly such that its density cannot be evaluated. This has driven the development of alternatives to the KL divergence such as the variational contrastive divergence \citep{ruiz2019contrastive}.
However, we are not aware of similar work for IS that leverages the computational efficiency of VI through a unifying loss such as the FKL divergence.

Other prior proposals combining IS and VI \citep{domke2018importance} have focused on minimizing the RKL, and thus do not inherently capture heavier tails of the target distribution. \citet{prangle2019distilling} recently presented concurrent work on combining IS and VI. However, their method relies on normalizing flows for constructing the proposal distribution such that it does not guarantee a multimodal approximation. \cite{ramos-lopez2018} uses a stream of IS weights to fit parameters for Gaussian mixture posteriors to minimize the FKL divergence. However, unlike our work, they assume access to normalized importance weights instead of samples.
\begin{table}[!ht]
\centering
\begin{tabular}{|c|c|c|c|c|c|}
\hline
{\footnotesize Failure Mode} & {\footnotesize VI} & {\footnotesize IS} & {\footnotesize VB} & {\footnotesize AIS} & {\footnotesize Ours} \\
\hline
{\footnotesize Multimodality}&\xmark& \xmark & \cmark & \cmark & \cmark \\
\hline
{\footnotesize Heavy tails}&\xmark& \cmark & \xmark & \cmark & \cmark \\
\hline
{\footnotesize Cov estimation}&\xmark& \cmark & \xmark & \cmark & \cmark \\
\hline
{\footnotesize High dimensions}&\cmark& \xmark & \cmark & \xmark & \cmark* \\
\hline
\end{tabular}
\caption{\label{tab:comp}Comparing approximate inference techniques.}
\end{table}
\section{Methodology}
We develop our novel approach to integrate variational inference and importance sampling using the forward KL divergence with a focus on multimodal targets.

Note that we assume the target density $p$ and the approximation $q$ share the same support which can be $\mathbb{R}^d$ or a subset thereof. This guarantees that $p$ is absolutely-continuous with respect to $q$ (noted as $p\ll q$) and vice-versa which is necessary for the definition of the KL divergence. 
%We explore this combination further in the multimodal setting leading to the formulation of a novel adaptive importance sampling with a mixture proposal distribution iteratively constructed by forward KL refinement. 
% based on the basic observation that the fewer and shallower the modes in distribution, the easier it is to sample ?

% \subsection{FKL variational approximation as a Proposal Distribution}
\subsection{Forward KL Variational Approximation for IS}
A theoretically reasonable desire is for the proposal distribution to minimize the forward KL divergence. However, we cannot compute FKL exactly for unnormalized target distributions. This stems from the expectation under the target $p$ in Eq.~\eqref{eq:fkl_decomp}. 

By contrast, the reverse KL takes the expectation in Eq.~\eqref{eq:rkl}) under a normalized approximation $q$. 
This can be seen as a tractable approximation to FKL with unknown bias~\citep{yao2018yes}. In fact, for misspecified choices of $\mathcal{Q}$ (i.e. when $\KL(q^*||p) \neq 0$), \cite{campbell2019universal} demonstrate that minimizing the RKL is not guaranteed to minimize the FKL divergence.
Furthermore, the light tails of the RKL minimization solution renders it inadequate for IS.

Alternatively, towards deriving a consistent approximation of the forward KL divergence, we rearrange densities inside the expectation as follows: 
\begin{equation}
    \KL(p\|q) = \E_p\left[\log{\left(\frac{p}{q}\right)}\right] = \E_{q}\left[\frac{p}{q} \log{\left(\frac{p}{q}\right)}\right].
\label{eq:fkl_rearrange}
\end{equation}
FKL can then be approximated through self-normalized importance sampling (SNIS) \citep{murphy2012machine} which is known to be consistent:
\begin{align}\label{eq:fkl_snis}
    \theta_s \sim q(\theta) \quad r_s = \frac{p(\theta_s|x)}{q(\theta_s)} \quad w_s = \frac{r_s}{\sum_{s=1}^{S} r_s} \nonumber\\
    \KL(p\|q) = \sum_{s=1}^{S} w_{s} \cdot \log{\left(\frac{p(\theta_s|x)}{q(\theta_s)}\right)}.
\end{align}
The SNIS estimation can have arbitrarily high variance depending on $q$. Gradients of Eq.~\eqref{eq:fkl_rearrange} with respect to the parameters of $q_i$ can also have high variance since we optimize over the same distribution from which samples are drawn. 
For certain distributions, the reparametrization trick can reduce this variance \citep{kingma2013auto}. Furthermore, the sequential setting described in section \ref{sec:lower_var} contributes to reducing the gradient variance. % Regardless, gradient variance is less of a concern in the following sequential approximation framework.\red{<rewrite}
\subsection{Forward KL Boosting}
\label{sec:fkl_boosting}
We sequentially construct a proposal mixture distribution $q$ that is both easy to sample from (lower SNIS bias) and minimizes the FKL objective for efficient IS estimation. 

Given the computational convenience of Gaussian distributions, we set $\mathcal{Q}$ to be the family of Gaussian mixtures in Section \ref{sec:experiments} such that the proposal at the $K^{th}$ boosting iteration is $q_K(\theta) := \sum_{i = 1}^K \lambda_i f_i(\theta; \phi_i)$ where $f_i = \mathcal{N}(\mu_i, \Sigma_i)$.

Our framework also supports mixtures of heavy-tailed distributions which are desirable for adaptive IS \citep{geweke1989bayesian}.

% The design of a variational family $\mathcal{Q}$ can be challenging a priori which often leads to model misspecification in VI estimates. Our experiments in Appendix~ \ref{sec:experiments} rely on the family of Gaussian mixtures due to the low variance of reparametrization gradients \citep{kingma2013auto}. 
% Gaussian mixtures can also arbitrarily approximate any absolutely continuous density \citep{epanechnikov1969non}. However, one drawback with Gaussians as mixture components is that there can be significant mismatch with a heavy tailed target distribution. IS can mitigate the effects of that mismatch; therefore, misspecification of the variational family is not a major concern for our approach. 

% Our proposal distribution at the $K^{th}$ boosting iteration is $q_K(\theta) := \sum_{i = 1}^K \lambda_i f_i(\theta; \phi_i)$. %In Section \ref{sec:experiments}, $f_i$ is the probability distribution function of the normal distribution, $\mathcal{N}(\mu_i, \Sigma_i)$.
% \begin{equation}
% q_k(\theta) :=
%     \sum_{i = 1}^k \lambda_i \mathcal{N}(\theta| \mu_i, \Sigma_i) \\
% \label{eq:mixture}
% \end{equation}

At each boosting iteration, we minimize the FKL between the mixture $q_i$ and the target $p$ by fitting a new component $f_i$ and a mixture weight $\lambda_i = \gamma$ while holding the parameters of previously-learned mixture components fixed:
\begin{equation*}\label{eq:mixture_forward_kl} 
        \underset{q_i \in \mathcal{Q}}{\argmin} \; \KL(p\|q_i) = \underset{f_i, \gamma}{\argmin} \; \KL\left(p\|\gamma f_i + (1 - \gamma)  q_{i-1}\right)
\end{equation*}
This is known to be more efficient and stable than the joint optimization of all mixture components at each iteration \citep{locatello2018boosting}. The mixture weights for the fixed mixture components $\{\lambda_j\}_{j=1}^{i-1}$ are re-scaled by $(1 - \gamma)$ and can be further adjusted with a fully-corrective weight search~\citep{locatello2018} using the gradient derived in Eq.~\eqref{eq:weight_grad}. Fitting the mixture weights is a convex problem that can benefit from higher-order optimizers such as Newton's.
% However, intuitively, the fewer and shallower the modes in a distribution, the easier it is to sample which we conjecture could improve the efficiency of importance sampling. (\red{is there no stronger motivation? it clearly does not align with Chatterjee's bound unless we can show that minimizing the FKL could be just as good as a diffuse proposal?})
We provide an alternative approach to building $q_K$ using the remainder distribution in Appendix~ \ref{app:fkl_boosting_alternate}.
\subsection{Lower Variance SNIS with Boosting}
\label{sec:lower_var}
The main computational concern in estimating the FKL divergence with SNIS is the variance of importance sampling in Eq.~\eqref{eq:fkl_rearrange}. However, the sequential construction described above enables further reformulations of the SNIS approximation which can lead to a lower variance than the naive approximation of Eq.~\eqref{eq:fkl_rearrange}.
In fact, the global objective can be re-written as:
\begin{align}
    &\underset{q_i \in \mathcal{Q}}{\argmin} \quad \KL(p \| q_i) = \underset{q_i \in \mathcal{Q}}{\argmin}\quad \E_p \left[\log \frac{p}{q_i}\right] \\
    &= \underset{f_i, \gamma}{\argmin}\quad \E_{q_i} \left[\frac{p}{q_{i}} \log{\frac{p}{\gamma f_i + (1-\gamma) q_{i-1}}}\right] \label{eq:new_mixture} \\ 
    &= \underset{f_i, \gamma}{\argmin}\quad \E_{q_{i-1}} \left[\frac{p}{q_{i-1}} \log{\frac{p}{\gamma f_i + (1-\gamma) q_{i-1}}}\right] \label{eq:old_mixture}
\end{align}

Eq.~\eqref{eq:old_mixture}, which is only possible in a sequential setting such as ours, reduces the gradient variance since the component being estimated, $f_i$, is independent of the distribution  $q_{i-1}$ from which samples are drawn to approximate the FKL. Furthermore, we can draw a large number of samples from $q_{i-1}$ a single time at the start of each boosting iteration.

The SNIS approximation of Eq.~\eqref{eq:old_mixture} given samples $\theta_s$ from $q_{i-1}$ can then be computed as:
\begin{equation*}\label{eq:old_mixture_snis}
     \sum_{s=1}^S w_s [\log{p(\theta_s)} - \log{(\lambda f_i(\theta_s) + (1-\lambda) q_{i-1}(\theta_s))}],
\end{equation*}
% \begin{equation}
%      \sum_{s=1}^S w_{\text{norm}}^s [\log{p(\theta_s)} - \log{(\lambda f_i(\theta_s) + (1-\lambda) q_{i-1}(\theta_s))}],
% \label{eq:old_mixture_snis}
% \end{equation}
where $w_s$ are computed as in Eq.~\eqref{eq:fkl_snis}. % the self-normalized importance weights 

\paragraph{Trade-off of sampling $q_i$ or $q_{i-1}$:} While $q_i$ is expected to handle well-separated modes better, in our experiments (\ref{sec:experiments} we found that using $q_{i-1}$ is sufficient for capturing distant modes (e.g. Fig.~\ref{fig:20mog}) while reducing the SNIS variance. Variance reduction was especially crucial on higher dimensional applications such as Bayesian NNs in Table.~\ref{tab:bnn_uci}.

% The remainder-based objective can also be re-written:
% $$\underset{f_i}{\argmin} \quad \KL(r_i \| f_i) = \underset{f_i}{\argmin} \quad \E_{f_i} \left[\frac{r_i}{f_i}\log{\frac{r_i}{f_i}}\right].$$
% % {\footnotesize
% % \begin{align*}
% %     &\underset{f_i}{\argmin} \quad \KL(r_i \| f_i) =\underset{f_i}{\argmin} \quad \E_{r_i} \left[\log{\frac{r_i}{f_i}}\right] \\
% %     &= \underset{f_i}{\argmin} \quad \E_{f_i} \left[\frac{r_i}{f_i}\log{\frac{r_i}{f_i}}\right].
% % \end{align*}
% % }
% % $\underset{f_i}{\argmin} \quad \KL(r_i \| f_i) =\underset{f_i}{\argmin} \quad \E_{r_i} \left[\log{\frac{r_i}{f_i}}\right] = \underset{f_i}{\argmin} \quad \E_{f_i} \left[\frac{r_i}{f_i}\log{\frac{r_i}{f_i}}\right].$

% This objective also enjoys lower gradient variance since the distribution being sampled from is a single Gaussian component and not a multimodal proposal. However, the light tails of a single Gaussian component can lead to high variance SNIS estimates in the case of severe mismatch with the target distribution.
% %
% Therefore, the choice of either should be guided by assumptions about the tails of the target.
% %
% However, given comparable empirical results of the two approaches, we limit the analysis of Section \ref{sec:analysis} and the experiments of Section \ref{sec:experiments} to the first and more straightforward approach.
%
% \subsection{Initialize with RKL, Refine with FKL}
% This sequential construction of the proposal can reduce the SNIS variance by drawing samples from the previously-estimated and fixed mixture distribution $q_{i-1}$. 
\subsection{Initialize with RKL, refine with FKL}
\label{sec:rkl_init}
For the first boosting iteration, we do not have an existing approximation to sample from. Instead, Eq.~\eqref{eq:old_mixture} can be re-written as follows:
\begin{equation*}
    \underset{f_i, \gamma}{\argmin} \; \E_{f_i} \left[\frac{p}{f_{i}} \log{\frac{p}{\gamma f_i + (1-\gamma) q_{i-1}}}\right].
\end{equation*}
% OLD VERSION 
% \begin{align*}
%     \underset{q_i \in \mathcal{Q}}{\argmin} \; \E_p \left[\log \frac{p}{q_i}\right]
%     = \underset{f_i, \gamma}{\argmin} \; \E_{f_i} \left[\frac{p}{f_{i}} \log{\frac{p}{\gamma f_i + (1-\gamma) q_{i-1}}}\right].
% \end{align*}
This exacerbates the sensitivity to the initialization of $f_i$. In fact, a sharply peaked initialization centered could limit the SNIS estimate from properly capturing the tail behavior. Therefore, a diffuse initialization is likely to be beneficial.

An even more practical initialization would use RKL-based VI to identify the mode of the target and provide a computationally efficient approximation that can be refined by FKL boosting in later iterations.
As such, this FKL-boosting workflow can be considered as general framework for the iterative refinement of any given posterior approximation to be used for the estimation of expectations of interest. This best combines the strengths of VI and IS as the first iteration of RKL minimization can reduce the variance of the SNIS approximation of FKL for the following iterations.% \red{<rewrite}
\section{Analysis}\label{sec:analysis}
% \red{}
We provide theoretical analysis of the proposed method of performing importance sampling with a proposal distribution constructed from FKL-based boosting.
\subsection{FKL Controls Moment Estimation Error}
Minimizing the FKL implicitly minimizes the error in posterior probabilities and moments via its control on total variational (through Pinsker's inequality\citep{tsybakov2008introduction}) and $l$-Wasserstein as follows~\citep{bolley2005}: Assume that the probability density $q$ is $n$-exponentially integrable. Then for target distribution $p$ such that $p \ll q$:
\begin{equation}\label{eq:Wasserstein_bound} 
W_n(q, p) \leq C_{n}^{EI} (q) \left[ \KL(p\|q)^{\frac{1}{n}} + \frac{\KL(p\|q)}{2}^{\frac{1}{2n}} \right],
\end{equation}
where 
$$C_{n}^{EI} (q) = \inf_{x_0\in\mathbb{R}^d, \alpha>0} \left( \frac{3}{\alpha} + \frac{2}{\alpha} \log\int e^{\alpha \|x-x_0\|^n} q(x) \rd x \right).$$
This implies the convergence of the first $n$ moments.

Note that because of the symmetry of Wasserstein distances, we can switch probability densities $p$ and $q$ in the above inequalities and obtain bounds in terms of RKL.
However, that would incur the $n$-exponential integrability condition on the target probability $p$, which boils down to generalized sub-gaussianity of its tail.% \red{<rewrite or drop}

The $n$-exponential integrability assumption is not required of the target density $p$ in the case of Eq.~\eqref{eq:Wasserstein_bound}. Instead, it is only required of the family of variational approximations $q$ which is easier to enforce and verify (and is automatically satisfied by the mixture of Gaussians). A smaller constant $C_n^{EI}$ can also be achieved by the same reasoning.
\subsection{FKL Boosting Converges at $O(1/K)$}
\label{sec:boosting_convergence}
Assuming $p \ll q$, which can be verified by the design of $\mathcal{Q}$, the functional gradient of the forward KL divergence is derived in Appendix~ \ref{sec:functional_gradient} as $
\frac{\delta KL(p||q)}{\delta q} = -\frac{p}{q}$.
The convexity of $\KL(p\|q)$ in $q$ is well established in the literature (proven with the log-sum inequality). Furthermore, we show in Appendix~ \ref{sec:boosting_convergence} that the FKL functional is also $\beta$-smooth in $q$ where $\beta$ depends on the range of the values that the density $q$ can take. This requires bounding $q$ away from zero and from above which is typical in prior theoretical work \citep{guo2016boosting} and aligns with practice as it can translate to a bounded parameter space for a given family of distributions.

For a convex and strongly smooth functional, the greedy sequential approximation framework of~\cite{zhang2003sequential}  provides an asymptotic guarantee for the convergence to a target distribution in the convex hull of a given base family such that the approximation error at the $K^{th}$ iteration is
$
    \KL(p \| q_K) = \KL \left(p \bigg\| \sum_{i=1}^K \lambda_i f_i\right) = O(1/K).
$
%a rate of $O(1/K)$ where $K$ is the number of boosting iterations.
%
This framework does not require each iteration to exactly solve for the optimal mixture component which can be difficult for the non-convex optimization sub-problems. 
% \subsection{Sample Complexity of SNIS: !!!REMOVE!!!}
% By Theorem~$1.1$ of~\citep{chatterjee2018sample}, we know that the variance of an importance sampling estimate scales as $O\left( \frac{e^{\KL(p\|q)}}{\sqrt{S}} \right),$
% where the number of samples $S$ required for importance sampling to provide accurate mean estimates scales exponentially with the KL divergence $\KL(p\|q)$.

\subsection{Computation-Quality Trade-Off}
While our iterative algorithm is guaranteed to converge asymptotically to the optimal proposal distribution, we can identify three sources of approximation error: the variational inference error, the SNIS approximation bias, and the greedy sequential approximation error which depend on the number of VI iterations, IS samples, and boosting iterations, respectively. 
These three errors
%hyperparameters 
thus finely control the compute-quality trade-off of our framework. We analyze these tradeoffs in more depth through simulation experiments below.

\subsection{The dependence of the IS proposal on the integrand}
Similarly to prior empirical \citep{owen2013monte} and theoretical \citep{chatterjee2018sample} works, we do not address any assumptions about the function $f$ being integrated. However, the optimal proposal $q^* \propto |f|\cdot p$ when $p$ is normalizable and $q^* \propto |f - I|\cdot p$ otherwise \citep{kahn1953}. 
Nonetheless, we only require that the integrand $f$ does not contain any singularities and shares the same support as $p$ and $q$. In this case, a simple rearrangement of the terms inside the expectation (\ref{eq:expectation}) would imply that $q$ should approximate $f\cdot p$ instead of $p$ where $f$ is the integrand. Therefore, characteristics of $f$ have no effect on our asymptotic guarantees or general methodology. %Accordingly, our decision to not explicitly discuss the integrands is aligned with prior work, both empirical \citep{owen2013monte} and theoretical \citep{chatterjee2018sample}. More generally, optimizing the FKL between $p$ and $q$ is reasonable if there are multiple functions $f$ or $f$ is unknown at the time of sampling.

\section{Simulation Experiments}
\label{sec:sim}
Using two illustrative simulation experiments, we provide further intuition for the behavior and performance of the proposed methodology using the FKL.

\subsection{Simulation 1: Cauchy}

\begin{figure*}[!ht]
    \centering
    \includegraphics[width=0.9\linewidth]{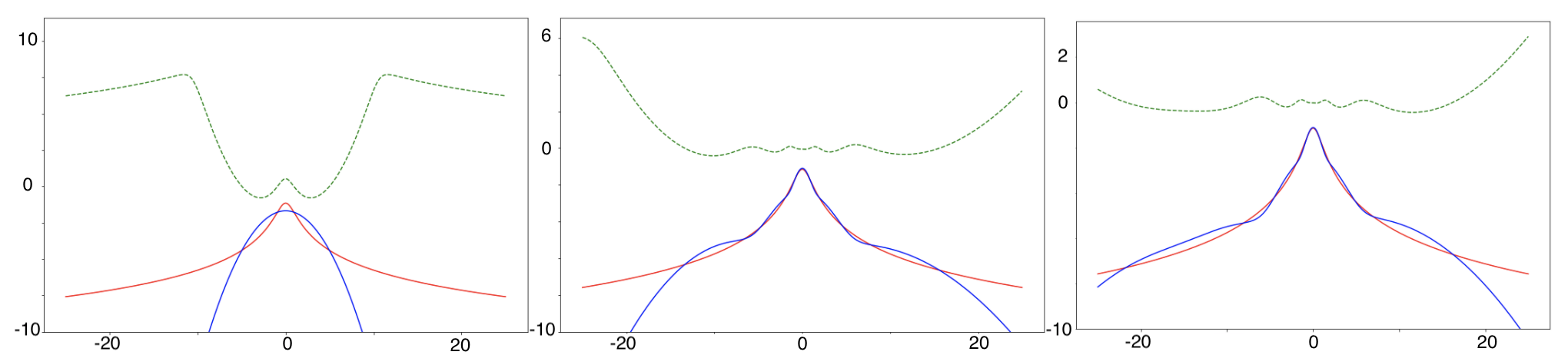}
    \caption{Log density plots of the result \blue{$\log q_k$ (blue, solid)}  of FKL boosting on the \red{Cauchy target ($\log p$) (red, solid)} . The \green{log-residual ($\log p/q$)(green, dashed) }  is indicative of the IS variance (from left to right: boosting iterations 1, 5, and 10).}
    \label{fig:cauchy_logpdf}
\end{figure*}

First, we demonstrate the aforementioned computation-quality trade-offs using a standard Cauchy target distribution. For intuition on the behavior of boosting using the FKL, Figure \ref{fig:cauchy_logpdf} illustrates the density plots of the target $p$ and the boosting approximation $q$ after various iterations.

\begin{figure}[!ht]
\centering
\includegraphics[width=0.9\linewidth]{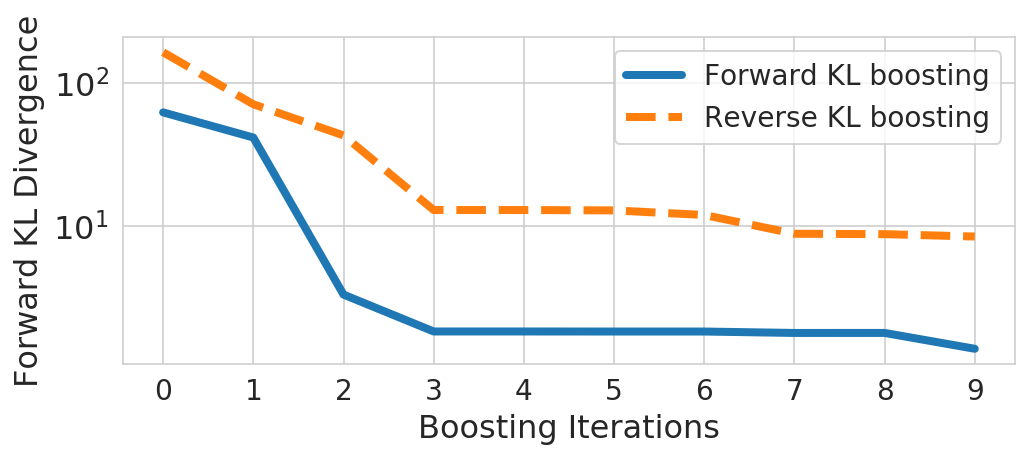}
\caption{A comparison in terms of the FKL divergence to a Cauchy target distribution over the course of variational boosting using the \blue{FKL} and \orange{RKL} divergences. } \label{fig:boosting_benefit}
\end{figure}

From previous sections, we observe a trade-off between sample complexity of SNIS and the optimization complexity of variational boosting.
%
% \paragraph{Optimization Complexity of Variational Boosting}
% On the other hand, when we use variational boosting to optimize $\KL(p\|q)$, we show in \ref{sec:boosting_convergence}, similarly to \citep{locatello2018} for the case of reverse KL, that for a fixed number $K$ of mixture components, if $q_K$ minimizes $KL(p\|q_K)$, then $KL(p\|q_K) = O(1/K)$.
%
Inclusion of more mixture components $K$ and more accurate optimization in the variational boosting steps can save exponentially many samples in SNIS. However, there is a diminishing gain in increasing $K$. 
% On top of that, the more mixture components we include, the harder it is to reach the global minimum of the variational objective.
We demonstrate in Fig.~\ref{fig:boosting_benefit} this effect: both FKL boosting and RKL boosting decreases forward KL divergence as more variational components are added. 
The decrease slows down significantly after inclusion of $3$ mixture components.
We therefore introduce in the experiments up to $3$ mixture components and select the best performance on validation data set.
From Fig.~\ref{fig:boosting_benefit} and the experimental results, we observe uniform improvements of FKL over RKL methods.

\subsection{Simulation 2: well-separated modes}
We next demonstrate the performance of boosting with the FKL on a distribution with a large number of well-separated modes. We set as the target distribution a 2-dimensional Gaussian mixture model (GMM) with 20 components, previously used by~\citet{ma2019irreversible}.

\begin{figure*}[!ht]
    \centering
    \includegraphics[width=\linewidth]{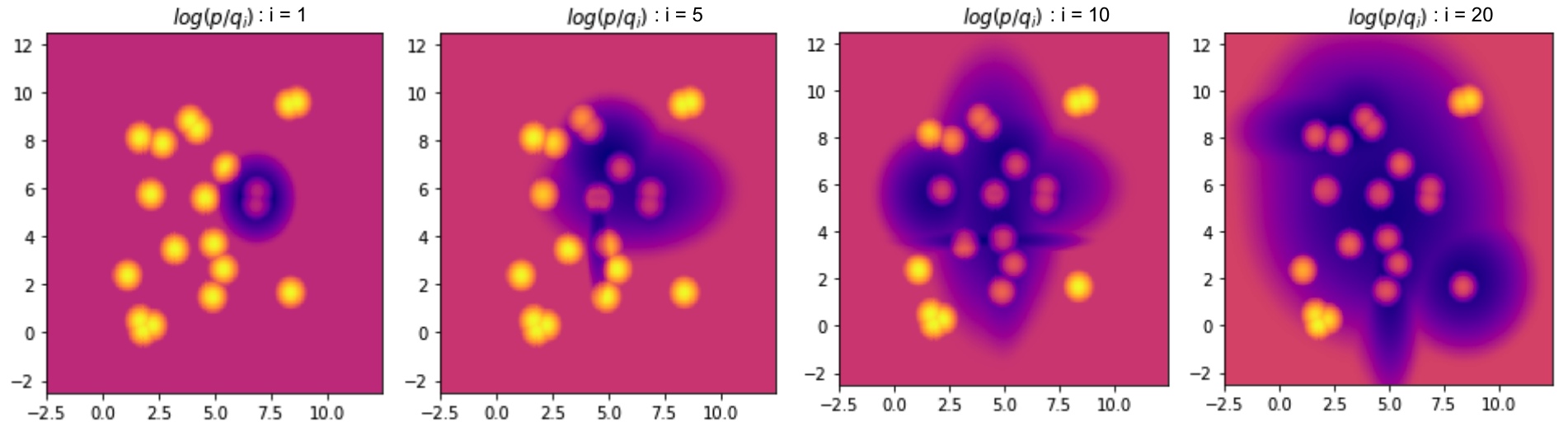}
    \caption{Log residual ($\log p/q_k$) plots of for FKL boosting on a 2-dimensional GMM of 20 components~\citep{ma2019irreversible}.}
    \label{fig:20mog}
\end{figure*}

The log-residual in Fig.~\ref{fig:20mog} demonstrates the sequential improvement of our approximation. Moment estimation results for this experiment can be found in  Fig.~\ref{fig:20mog_res} in the Appendix. Overall, boosting with the FKL effectively improves both moment estimation and the actual FKL, and outperforms the RKL as the number of boosting iterations increases.
\section{Real Data Experiments}\label{sec:experiments}
We evaluate the performance of the proposed method when applied to Bayesian linear regression (BLR) and Bayesian neural networks (BNNs) using a Gaussian prior and a heavy tailed prior. We use four datasets from UCI \citep{UCI} (Table \ref{tab:datasets} in Appendix~ \ref{app:experiments}). We split each dataset into twenty randomly drawn 90\%/10\% train/test splits, which we denote $\mathcal{D}_{\text{train}} = \{x_i, y_i\}_{i=1}^{N_{\text{train}}}$ and $\mathcal{D}_{\text{test}} = \{x_i, y_i\}_{i=1}^{N_{\text{test}}}$, with input $x_i$ and output $y_i$. We report the mean and std. dev. of results over all splits.

We demonstrate our proposed method with $K=$ 1, 2, and 3. We refer to fitting a single Gaussian, or $K=1$, as FKL VI. For runs with more than one boosting iteration, we initialize the first component using the RKL, and optimize subsequent iterations using the FKL (as described in Section \ref{sec:rkl_init}). 
% The reason for this is that in practice, we found that the forward KL leads to a more diffuse approximation in the first iteration which can make initializing the next component challenging as the first component can cover most of the distribution mass leaving no residual to guide the mode initialization of following components. Using RKL for the first iteration, however, allows to identify the most prominent mode in the target distribution but the approximation tends to be light-tailed which facilitate initialization based on the residual. We believe this is a limitation of our initialization heuristic and not our overall approach.

% We fit the proposal $q$ directly to the target. Fitting the proposal $q$ to the remainder led to comparable results, which we report in Appendix \ref{app:experiments}.

\textbf{Optimization details:} We use the ADAM optimizer \citep{kingma2014adam} for each boosting iteration with a fixed learning rate and compute gradients based on a fixed number of samples using Autograd \citep{maclaurin2015}. At the end of each boosting iteration, mixture weights are fully re-optimized using simplex-projected gradient descent \citep{bubeck2014convex} based on the analytical gradient in Eq.~\eqref{eq:weight_grad}. Details about practical considerations and hyperparameters can be found in Appendix~\ref{app:experiments}, and we include our code with the submission.

\textbf{Parameter initialization:}
We follow standard practice with similar experiments (see e.g. \citep{miller2017}) where the means of each component are initialized at zero and the diagonal elements of the initial covariance matrix are drawn from $\mathcal{N}(0, \sigma^2)$ ($\sigma$ is tuned as a hyperparameter, see Appendix~~\ref{app:hparams}). At the start of each boosting iteration, we further apply an initialization heuristic which approximates the mode of the residual by gradient descent on the remainder density. This requires a single sample per gradient step and is run for 400 steps.
% which is equivalent to two FKL minimization iterations in our experiments. 
Note that this initialization is significantly simpler and less compute-intensive than prior approaches which include a weighted EM algorithm \citep{miller2017} and a random search over $10,000$ samples or more \citep{campbell2019universal}.

% \textbf{Parameterization:} The covariance matrices for each component were parameterized to be diagonal and PSD. In each boosting iteration, all mixture weights were fully re-optimized using simplex-projected gradient descent and Eq.~\eqref{eq:weight_grad}. Further details can be found in Appendix~ \ref{app:experiments}, and we include our code with the submission.

\textbf{Comparisons:}
We compare our approach to variational inference (RKL VI) and variational boosting with 2 and 3 boosting iterations (RKL VB) \citep{miller2017}. For the comparison to RKL VI and RKL VB, we use the same parametrization, initialization, and optimization techniques as for FKL VI and FKL VB. This might lead to discrepancies compared to the published results \citep{miller2017}; however, keeping these details consistent better disentangles the effect of the RKL vs. FKL optimization. 

We also compare to directly sampling from the target distribution using Hamiltonian Monte Carlo (HMC) \citep{neal2011hmc}, implemented using the TensorFlow Probability library \citep{tfp}. We additionally ran 3 HMC chains in parallel and averaged the results, similar to \citep{hoffman2020}. Results were comparable between 1 and 3 HMC chains, and we include the results for 3 chains in Appendix~ \ref{app:experiments}. 
% \subsection{Synthetic example: Heavy-Tailed Parameter Inference}
% \red{TODO: if no toy results, drop this section}

% Given that our final goal is to best approximate Eq. \ref{eq:expectation}, we can evaluate the difference of expectations of a range of functions computed based on samples from the ground-truth normalized target distributions (only available in synthetic settings, typically) the importance sampling proposal:
% \begin{align}
%     Diff_{VB}(p, q) &= \|E_p[f] - E_q[f]\|^2\\
%     Diff_{IW}(p, q) &= \|E_p[f] - E_q[p/q f]\|^2
% \end{align}
% This can be assessed for specific classes of functions which can reduce to known metrics or simply for functions such as $X$, $X^2$ which also reduces to the well known $W_1$ and $W_2$ metrics.   
%%%%%%%%% NEW TABLE %%%%%%%%%%
\begin{table*}[!ht]
\centering
\begin{tabular}{|c|c|c|c|c|}
\hline
Method & Wine {\footnotesize ($d=14$)} & Boston {\footnotesize ($d=16$)}& Concrete {\footnotesize ($d=11$)} & Power {\footnotesize ($d=7$)} \\
\hline
{\footnotesize HMC} & -1.002	{\footnotesize ($\pm$ 0.012)}& -2.923	{\footnotesize ($\pm$ 0.035)}& -3.780	{\footnotesize ($\pm$ 0.013)}& -2.923	{\footnotesize ($\pm$ 0.006)} \\
\hline
{\footnotesize RKL VI} & -1.013	{\footnotesize ($\pm$ 0.013)}& -2.947	{\footnotesize ($\pm$ 0.035)}& -3.798	{\footnotesize ($\pm$ 0.013)}& -2.921	{\footnotesize ($\pm$ 0.006)} \\
{\footnotesize RKL VB 2} & -1.014	{\footnotesize ($\pm$ 0.011)}& -2.924	{\footnotesize ($\pm$ 0.033)}& -3.787	{\footnotesize ($\pm$ 0.014)}& -3.067	{\footnotesize ($\pm$ 0.003)} \\
{\footnotesize RKL VB 3} & -1.007	{\footnotesize ($\pm$ 0.012)}& -2.945	{\footnotesize ($\pm$ 0.039)}& -3.800	{\footnotesize ($\pm$ 0.015)}& -3.073	{\footnotesize ($\pm$ 0.003)} \\
\hline
{\footnotesize FKL VI} & \textbf{-0.998}	{\footnotesize ($\pm$ 0.012)}& -2.905	{\footnotesize ($\pm$ 0.036)}& -3.775	{\footnotesize ($\pm$ 0.014)}& \textbf{-2.920}	{\footnotesize ($\pm$ 0.005)} \\
{\footnotesize FKL VB 2} & \textbf{-0.998}	{\footnotesize ($\pm$ 0.012)}& -2.906	{\footnotesize ($\pm$ 0.036)}& \textbf{-3.762}	{\footnotesize ($\pm$ 0.013)}& -2.921	{\footnotesize ($\pm$ 0.006)} \\
{\footnotesize FKL VB 3} & \textbf{-0.998}	{\footnotesize ($\pm$ 0.013)}& \textbf{-2.904}	{\footnotesize ($\pm$ 0.036)}& \textbf{-3.762}	{\footnotesize ($\pm$ 0.013)}& -2.921 {\footnotesize ($\pm$ 0.006)} \\
\hline
\end{tabular}
\caption{\label{tab:blr_uci}Predictive log probabilities on test for BLR with Gaussian prior (mean $\pm$ standard error over 20 train/test splits).}
\end{table*}
\subsection{BLR and BNNs with Gaussian Priors}\label{sec:blr_gaussian}
\begin{table*}[!ht]
\centering
\begin{tabular}{|c|c|c|c|c|c|c|}
\hline
Method & Wine {\footnotesize ($d=653$)}& Boston {\footnotesize ($d=753$)} & Concrete {\footnotesize($d=503$)} & Power {\footnotesize($d=303$)} \\
\hline
{\footnotesize HMC} & -0.990	{\footnotesize ($\pm$ 0.014)}& -2.709	{\footnotesize ($\pm$ 0.101)}& -3.281	{\footnotesize ($\pm$ 0.017)}& \textbf{-2.817}	{\footnotesize ($\pm$ 0.007)} \\
\hline
{\footnotesize RKL VI} & -0.993	{\footnotesize ($\pm$ 0.014)}& -2.858	{\footnotesize ($\pm$ 0.020)}& -3.230	{\footnotesize ($\pm$ 0.015)}& -2.851	{\footnotesize ($\pm$ 0.008)} \\
{\footnotesize RKL VB 2} & -0.990	{\footnotesize ($\pm$ 0.015)}& -2.832	{\footnotesize ($\pm$ 0.018)}& -3.253	{\footnotesize ($\pm$ 0.015)}& -2.945	{\footnotesize ($\pm$ 0.009)} \\
{\footnotesize RKL VB 3} & -0.981	{\footnotesize ($\pm$ 0.012)}& -2.744	{\footnotesize ($\pm$ 0.011)}& -3.255	{\footnotesize ($\pm$ 0.017)}& -3.002	{\footnotesize ($\pm$ 0.012)} \\
\hline
{\footnotesize FKL VI} & -0.991	{\footnotesize ($\pm$ 0.015)}& \textbf{-2.677}	{\footnotesize ($\pm$ 0.011)}& 	-3.328	{\footnotesize ($\pm$ 0.019)}& -2.872	{\footnotesize ($\pm$ 0.012)} \\
{\footnotesize FKL VB 2} & -0.979	{\footnotesize ($\pm$ 0.017)}& -2.779	{\footnotesize ($\pm$ 0.012)}& -3.193	{\footnotesize ($\pm$ 0.016)}& -2.870	{\footnotesize ($\pm$ 0.008)} \\
{\footnotesize FKL VB 3} & \textbf{-0.967}	{\footnotesize ($\pm$ 0.014)}& -2.801	{\footnotesize ($\pm$ 0.012)}& \textbf{-3.192}	{\footnotesize ($\pm$ 0.016)}& -2.851	{\footnotesize ($\pm$ 0.009)} \\
\hline
\end{tabular}
\caption{\label{tab:bnn_uci}Predictive log probabilities on test for BNNs with Gaussian prior (mean $\pm$ standard error over 20 train/test splits).}
\end{table*}

We follow the experimental setup of \citep{miller2017} for both BNNs and BLR. We place a Gaussian prior over each weight in the model, and an inverse Gamma prior on the variances:
\begin{align*}
    \alpha \sim \text{Gamma}(1, 0.1);  \quad \tau \sim \text{Gamma}(1, 0.1); \\
    w_i \sim \mathcal{N}(0, 1/\alpha); \quad  y | x, w, \tau \sim \mathcal{N}(\phi(x, w), 1/\tau),
\end{align*}
where $w$ is the set of weights, and $\phi(x,w)$ is either a linear function of $x$ (BLR) or a multi-layer perception (BNN). For our BNNs, we set $\phi$ to be a one-hidden layer neural network with 50 hidden units and ReLU activation function, as done by \citep{miller2017, hernandez2015}.
The full set of parameters that we sample is $\theta = (w, \alpha, \tau)$. We use the posterior predictive distribution to compute the distribution for a given new input $x$: 
\begin{equation}\label{eq:post_pred_dist}
    p(y | x, \mathcal{D}_{\text{train}}) = \int p(y | x, \theta) p(\theta | \mathcal{D}_{\text{train}}) d\theta.
\end{equation}

We use importance sampling to estimate this posterior predictive distribution given $S$ samples $\theta_s \sim q(\theta)$:
\begin{align}
   &p(y | x, \mathcal{D}_{\text{train}}) \approx \frac{1}{S} \sum_{i=1}^S \frac{p(\theta_s | \mathcal{D}_{\text{train}})}{q(\theta_s)} p(y | x, \theta_s),
\label{eq:iwll}
\end{align}
where $q(\theta)$ is the proposal distribution fit to $p(\theta | \mathcal{D}_{\text{train}})$ using either forward KL refinement (our method, FKL VB) or reverse KL refinement (RKL VB, \citep{miller2017}). 

Note that \citet{miller2017} does not use importance sampling to estimate the posterior predictive distribution. We add importance weights here as an ablation to limit our analysis to the difference between FKL and RKL optimization. For completness, we report the estimates without IS using RKL VB in Appendix~ \ref{app:experiments}. 
For comparing to HMC, we do not use importance sampling due to lack of an explicit density function, and instead compute (Eq.~\ref{eq:post_pred_dist}) by averaging over direct HMC samples from the posterior distribution $p(\theta | \mathcal{D}_{\text{train}})$ (Eq.~\eqref{eq:vbll} in Appendix~ \ref{app:experiments}). 

As our final evaluation metric, we report the average predictive log probabilities on held-out test data: 
\begin{equation}\label{eq:pred_log_prob}
    \frac{1}{|\mathcal{D}_{\text{test}}|}\sum_{x,y \in \mathcal{D}_{\text{test}}} \log p(y | x, \mathcal{D}_{\text{train}}).
\end{equation}
\begin{table*}[!ht]
\centering
\begin{tabular}{|c|c|c|c|c|}
\hline
Method & Wine {\footnotesize ($d=13$)} & Boston {\footnotesize ($d=15$)}& Concrete {\footnotesize ($d=10$)} & Power {\footnotesize ($d=6$)} \\
\hline
{\footnotesize HMC} & -1.004	{\footnotesize ($\pm$ 0.012)}& -2.962	{\footnotesize ($\pm$ 0.033
)}& -3.808	{\footnotesize ($\pm$ 0.026)}& \textbf{-2.916}	{\footnotesize ($\pm$ 0.005)} \\
\hline
{\footnotesize RKL VI} & -1.011	{\footnotesize ($\pm$ 0.013)}& -2.924	{\footnotesize ($\pm$ 0.037)}& -3.789	{\footnotesize ($\pm$ 0.013)}& -2.924	{\footnotesize ($\pm$ 0.006)} \\
{\footnotesize RKL VB 2} & -1.007	{\footnotesize ($\pm$ 0.014)}& -2.944	{\footnotesize ($\pm$ 0.031)}& -3.788	{\footnotesize ($\pm$ 0.012)}& -3.028	{\footnotesize ($\pm$ 0.003)} \\
{\footnotesize RKL VB 3} & -1.008	{\footnotesize ($\pm$ 0.012)}& -2.940	{\footnotesize ($\pm$ 0.035)}& -3.796	{\footnotesize ($\pm$ 0.014)}& -3.005	{\footnotesize ($\pm$ 0.004)} \\
\hline
{\footnotesize FKL VI} & -0.993	{\footnotesize ($\pm$ 0.013)}& \textbf{-2.904}	{\footnotesize ($\pm$ 0.036)}& -3.775	{\footnotesize ($\pm$ 0.014)}& -2.940	{\footnotesize ($\pm$ 0.005)} \\
{\footnotesize FKL VB 2} & \textbf{-0.973}	{\footnotesize ($\pm$ 0.009)}& -2.907	{\footnotesize ($\pm$ 0.035)}& \textbf{-3.773}	{\footnotesize ($\pm$ 0.015)}& -2.921	{\footnotesize ($\pm$ 0.006)} \\
{\footnotesize FKL VB 3} & -0.975	{\footnotesize ($\pm$ 0.009)}& -2.906	{\footnotesize ($\pm$ 0.036)}& -3.774	{\footnotesize ($\pm$ 0.015)}& -2.922	{\footnotesize ($\pm$ 0.006)} \\
\hline
\end{tabular}
\caption{\label{tab:heavy_blr_uci}Predictive log probabilities on test for BLR with heavy tailed prior (mean $\pm$ standard error over 20 train/test splits).}
\end{table*}
\subsection{BLR with Heavy Tailed Priors}\label{sec:blr_heavy}
In addition to the Gaussian prior, we also perform Bayesian linear regression with a heavy tailed prior. Following \citep{campbell2019universal} we place a $\mathcal{T}_2$ prior on the weights. We use the same inverse Gamma prior on the variance:
\begin{align*}
    \tau \sim \text{Gamma}(1, 0.1); \quad w \sim \mathcal{T}_2(0, A^TA); \\
    % b \sim \text{Cauchy}(0,1) \\
    y | x, w, \tau \sim \mathcal{N}(\phi(x, w), 1/\tau),
\end{align*}
where $A$ is fixed, and each entry is drawn i.i.d.\ before the optimization process: $A_{ij} \sim \mathcal{N}(0,1)$. For these BLR experiments, $\phi(x,w)$ is a linear function of $x$ with weight parameters $w$. The full set of parameters that we sample is $\theta = (w, \tau)$. We estimate the same posterior predictive distribution in Eq.~\eqref{eq:post_pred_dist} using IS in Eq.~\eqref{eq:iwll}, and report the average predictive log probabilities from Eq.~\eqref{eq:pred_log_prob}.

\section{Discussion}
\label{sec:discussion}
Tables \ref{tab:blr_uci}, \ref{tab:bnn_uci}, and \ref{tab:heavy_blr_uci} present the results  on the UCI datasets for BLR with a Gaussian prior, BNNs with a Gaussian prior, and BLR with a heavy tailed prior, respectively. The lowest mean predictive log probability is highlighted in bold.
%With the exception of the Power dataset, the highest average predictive log probabilities were achieved by either FKL VI or FKL VB. Interestingly, both FKL VB and RKL VB exhibited a deterioration in the held-out log-likelihood that can be attributed to overfitting to the training distribution which we discuss in Section \ref{sec:discussion}. 
% For the heavy tailed prior, the FKL methods achieved either the best or statistically similar average predictive log probabilities compared to HMC and RKL.

Minimizing the FKL divergence outperforms RKL across all four datasets and three experimental settings. This demonstrates the inadequacy of RKL-based VI for the construction of IS proposals. This should incentivize wider adoption of the FKL divergence, especially when the downstream task extends beyond simple prediction and requires a calibrated estimation of the posterior predictive distribution. FKL outperforming HMC, the gold standard of Bayesian inference, on three of the four datasets across all settings is another promising result.

However, we observe a decay in the held-out log-likelihood as more components are added for certain datasets (e.g. FKL VB with BNN on Boston). This is consistent with prior variational boosting results on the same datasets \citep{miller2017} and typically signals an over-fitting problem.
%look contradictory to the $O(1/K)$ rate. 
Therefore, it is worth emphasizing that the convergence analysis of Section \ref{sec:analysis} is limited to optimization guarantees and does not extend to learning or generalization guarantees. 
% As such, it is possible that our training error is decaying with each boosting iteration. 
% In fact, each iteration increases the dimensionality of our model which becomes more prone to overfitting on these small UCI datasets.

\paragraph{Computational considerations}
%One main advantage of RKL variational proposals argued in the literature is that they have relatively little computational overhead.
One main advantage of RKL methods is the low computational overhead, especially as compared to MCMC methods.
In our experiments we observed that, even in high dimensions, there do not seem to be significant computational differences between optimizing the FKL objective and optimizing the RKL objective.
In fact, our reported results compare RKL and FKL methods for the same number of IS samples and optimization iterations. See Appendix \ref{app:experiments} for exact hyperparameter values and wall clock times.

\section{Conclusion}

Overall, we propose a principled algorithm that combines the strengths of importance sampling and variational inference to efficiently approximate multimodal and possibly heavy-tailed targets. 
% In order to produce better estimates of an expectation over an intractable target distribution, 
% \red{To ensure a good proposal distribution for importance sampling, we iteratively refine a variational approximation to the target by minimizing the FKL divergence.}
Unlike prior work that relies on RKL, our minimization of FKL aligns with the analysis of the variance of IS~\citep{chatterjee2018sample} which guarantees an optimal proposal distribution asymptotically. 
%
% Empirically, our proposal indeed performs better than variational boosting on heavy tailed and high dimensional posterior distributions.
%
One challenge for this approach is the variance of the SNIS estimate of the forward KL divergence. Developing variance-reduction schemes for these types of objectives is an open research problem. Nonetheless, existing techniques for re-sampling or smoothing the importance weights can immediately apply to our proposed method.
% whose solutions will benefit the adoption of the FKL and related approaches.

% \clearpage
% \newpage

% \begin{contributions} % will be removed in pdf for initial submission,
%                       % so you can already fill it to test with the
%                       % ‘accepted’ class option
%     Briefly list author contributions.
%     This is a nice way of making clear who did what and to give proper credit.

%     H.~Q.~Bovik conceived the idea and wrote the paper.
%     Coauthor One created the code.
%     Coauthor Two created the figures.
% \end{contributions}

% \begin{acknowledgements} % will be removed in pdf for initial submission,
%                          % so you can already fill it to test with the
%                          % ‘accepted’ class option
%     Briefly acknowledge people and organizations here.

%     \emph{All} acknowledgements go in this section.
% \end{acknowledgements}

\newpage

\bibliography{references}

\newpage

\appendix

\section{Further methodology discussion}

\subsection{Combining HMC and FKL VB}
Using an MCMC method such as HMC can exploit the fact that the unnoramlized density $r_i$ gets shallower and less multimodal over iterations which makes it increasingly easy to sample from once, at the beginning of each boosting iteration, via techniques such as Hamiltonian Monte Carlo \citep{neal2011hmc}. While we saw some promising preliminary performance with this method on real data experiments using Bayesian logistic regression (BLR), the higher dimensional experiments with Bayesian neural networks (BNN) struggled with numerical instability that would require further tuning of the HMC hyperparameters which include the number of burn-in steps, learning rate, and number of leapfrog steps, among others.

% Note: we know from classical IS results that the optimal proposal $q^* \propto |f|*p$ or $q^* \propto |f - I|*p$. As such, the actual form of the function we're estimating the expectation of is not of great importance as we can easily re-define the target as $|f|*p$ or $|f - I|*p$ if needed. 

\subsection{Stabilization of likelihood ratios}
To avoid high-variance gradient estimates due to a mismatch between the target and the proposal, especially at the beginning of inference, we stabilize the importance weights in an unbiased way that parallels the log-sum-exp trick \citep{nielsen2016guaranteed} in order to handle potential under- or overflow when exponentiating large values:
\begin{align*}
    r_s &= \frac{p(\theta_s|x)}{q_i(\theta_s)}, \quad d_{max} = \underset{s}{\max}{(\log{p(\theta_s|x)} - \log{q_i(\theta_s)})} \\
    d_s &= exp\left(\log{p(\theta_s|x)} - \log{q_i(\theta_s)} - d_{max}\right) \\
    w_s &= \frac{d_s}{\sum_{s=1}^{S} d_s} = \frac{r_s}{\sum_{s=1}^{S} r_s}
\end{align*}

As for the log residual $\log{p/q}$ we introduce a biased stabilization heuristic that is typical in variational boosting \citep{guo2016boosting,campbell2019universal}
with an $\epsilon = e^{-10}$:
\begin{equation}
    \log{\left(\frac{p(\theta|x) + \epsilon}{q_i(\theta_s) + \epsilon}\right)} \approx  \log{\left(\frac{p(\theta|x)}{q_i(\theta_s)}\right)} 
\end{equation}

A range of variance reduction schemes is applicable to our method such a weight clipping, re-sampling and re-weighting. However, we leave that for future work.
% \begin{algorithm}[ht]
% \caption{Forward KL Refinement}\label{alg:boost}
% \begin{algorithmic}[1]
% \State \red{TODO: add details}
% \State \textbf{input:} $p$, the un-normalized target.
% \If{mixed = True}
%     \State $\mu_0, \Sigma_0 \gets \underset{\mu, \Sigma, \lambda}{\argmin} \; D_{KL}( \mathcal{N}(\theta; \mu, \Sigma) \|p(\theta))  $
% \Else
%     \State $\mu_0, \Sigma_0 \gets \underset{\mu, \Sigma, \lambda}{\argmin} \; D_{KL}(p(\theta) \|\mathcal{N}(\theta; \mu, \Sigma)) ) $
% \EndIf
% \State $\lambda_0 \gets 1$
% \For {$i = 2$ to $K$}
% \If{Approach A}
%     \State Draw samples from $q_{i-1}$
%     \State $\mu_i, \Sigma_i, \lambda_i \gets \underset{\mu, \Sigma, \lambda}{\argmin} \; D_{KL}( p(\theta) \|\lambda f_i(\theta; \mu, \Sigma) + (1 - \lambda) q_{i-1}(\theta) $
% \Else
%     \State $r_i(\theta) \gets \frac{p(\theta)}{q_{i-1}(\theta)}$
%     \State $\mu_i, \Sigma_i \gets \underset{\mu, \Sigma}{\argmin} \; D_{KL}(r_i(\theta) \| f_i(\theta; \mu, \Sigma)) $
% \EndIf
% \State $\{\lambda_i\}_{i=1}^{K} \gets FC(\{\mu_i\}_{i=1}^{K}, \{\Sigma_i\}_{i=1}^{K}, \{\lambda_i\}_{i=1}^{K})$ 
% \EndFor
% \State \Return $\{\mu_i\}_{i=1}^{K}$, $\{\Sigma_i\}_{i=1}^{K}$, $\{\lambda_i\}_{i=1}^{K}$
% \end{algorithmic}
% \end{algorithm}

\section{Derivations}
\label{sec:derivations}

\subsection{Connecting Forward KL to other metrics used for VI}
\label{sec:fkl_connections}
By the monotonicity of Renyi-$\alpha$ divergences, given that $\lim_{\alpha \to 1} D_{\alpha}(p,q) = \KL(p\|q)$, and from \citep{dieng2017variational} :
\begin{equation}
\KL(p\|q) \leq D_2(p,q) \leq \chi^2(p,q)
\end{equation}

\subsection{Reverse KL remainder: intrinsic entropy regularization}
\label{sec:remainder_entropy}
Computing the remainder-reverse KL objective using our approach leads to the well-known although usually ad-hoc entropy regularization (e.g. \citep{locatello2018boosting}).
\begin{align}
\KL(f_i \| r_i) &= \KL(f_i \| \frac{p}{q_{i-1}}) \\
&= \E_{f_i} [\log{\frac{f_i q_{i-1}}{p}}] \\
&= \E_{f_i} [\log{\frac{q_{i-1}}{p}}] + \E_{f_i}[\log{f_i}]
\end{align}

As we can see, while the first term is the mean of the log-residual under the new component $f_i$, the typical objective for gradient boosting, the second term is the entropy of $f_i$.
\subsection{SNIS derivation}
\label{sec:snis_derivation}
Since we do not assume to know the normalization constant of $p$, we shall approximate the above quantities by self-normalized importance sampling while making the distinction between the normalized $p$ and the un-normalized $\hat{p}$:
\begin{equation}
\theta_s \sim q_{i-1}, \quad w^s = \frac{\hat{p(\theta_s)}}{q_{i-1}(\theta_s)}, \quad w_{norm}^s = \frac{w^s}{\sum_s w^s}   
\end{equation}
\begin{align*}
    &\E_{q_{i-1}} [\frac{p}{q_{i-1}} \log \frac{p}{\lambda f_i + (1-\lambda) q_{i-1}}] \\ 
    &= \frac{\E_{q_{i-1}} [\frac{\hat{p}}{q_{i-1}} \log \frac{p}{\lambda f_i + (1-\lambda) q_{i-1}}]}{\E_{q_{i-1}} [\frac{\hat{p}}{q_{i-1}}]} \\
    &\approx \sum_s \frac{\frac{\hat{p(\theta_s)}}{q_{i-1}(\theta_s)}}{\sum_s \frac{\hat{p(\theta_s)}}{q_{i-1}(\theta_s))}} [\log \frac{p(\theta_s)}{\lambda f_i(\theta_s) + (1-\lambda) q_{i-1}(\theta_s)}] \\
    &= \sum_s w_{norm}^s [\log{p(\theta_s)} - \log{(\lambda f_i(\theta_s) + (1-\lambda) q_{i-1}(\theta_s))}]
\end{align*}

\subsection{Gradients of mixture weights}
For forward KL:
\label{sec:weight_grad}
\begin{align}
&\nabla_{\lambda_i} \E_p\left[ \log{p} - \log{\sum_j^K \lambda_j q_j}\right] = -\E_p \left[\nabla_{\lambda_i} \log{\sum_j^K \lambda_j q_j}\right] \nonumber \\
&= -\E_p\left[\frac{q_i}{\sum_j^K \lambda_j q_j}\right] = -\E_{q_i}\left[\frac{p}{q}\right]. \label{eq:weight_grad}
\end{align}

For reverse KL:
\begin{align*}
\nabla_{
\lambda_i} \E_q[ \log{q} - \log{p}] &= \E_{\sum_j^K \lambda_j q_j}[\nabla_{\lambda_i} \log{(\sum_j^K \lambda_j q_j)}] \\
&= \E_{q_i}[\log{q} - \log{p}]
\end{align*}

\subsection{The functional gradient of the forward KL divergence}
\label{sec:functional_gradient}
We assume $\supp p \subseteq \supp q$: that is, $p$ is absolutely continuous with respect to the variational approximation $q_i$ which can be ensured by the design of the variational family $\mathcal{Q}$.

Let $D(q) = \KL(p\|q)$. Functional gradient $\frac{\delta D}{\delta q}$ can be computed from the Taylor expansion of the KL functional \citep{friedman2001greedy} as follows:
\begin{equation}
    \lim_{\epsilon \to 0} \frac{D(q+\epsilon \cdot h) - D(q)}{\epsilon} = \int \frac{\partial D}{\partial q} h dx
\end{equation}
\begin{align*}
      \frac{D(q+\epsilon \cdot h) - D(q)}{\epsilon} &= \frac{1}{\epsilon} \int p \log{p} - p \log{(q + \epsilon h)} \\
      &+ p \log{p} - p \log{q} \\
      &= - \frac{1}{\epsilon} \int p \log{(q + \epsilon h)}  - p \log{q} \\
      &= - \frac{1}{\epsilon} \int p \log{(1 + \epsilon \frac{h}{q})}
\end{align*}
We have the logarithmic inequality $\frac{x}{x+1} \leq \log{(1+x)} \leq x \forall x > -1$ where we can substitute $\epsilon \frac{h}{q} > 0$ for x and arrive at 
\begin{equation*}
    -\frac{h}{q} \leq -\frac{1}{\epsilon} \log{(1 + \epsilon \frac{h}{q})} \leq - \frac{\frac{h}{q}}{1 + \epsilon \frac{h}{q}}
\end{equation*}
By the monotone convergence theorem we can take the limit inside the integral and arrive at 
\begin{align}
    \lim_{\epsilon \to 0} \int -p \frac{1}{\epsilon} \log{(1 + \epsilon \frac{h}{q})} = \int - p\frac{h}{q}
\end{align}
$$\frac{\delta D(q)}{\delta q} = -\frac{p}{q}$$

\subsection{Boosting convergence analysis}
For a convex and strongly smooth functional, the greedy sequential approximation framework of \citep{zhang2003sequential} provides an asymptotic guarantee for the convergence to a target distribution in the convex hull of the base family at a rate of $O(1/K)$ where $K$ is the number of boosting iterations.
This framework does not require each iteration to exactly solve for the optimal mixture component which can be difficult in variational inference. % however, the error sequences needs to converge to 0 in \infty.

While the convexity of $\KL(p\|q)$ in $q$ is well established in the literature (proven with the log-sum inequality) for the forward KL divergence functional, we can show that FKL is also $\beta$-smooth in $q$ where $\beta$ depends on the maximum and minimum values that the density $q$ can take.
To establish strong smoothness, on the other hand, stricter assumptions about the densities are necessary. If we assume that all densities are bounded away from 0 and from above $q_1$ then for any pair of densities $q_1$ and $q_2$ there exists a $\beta = \sup \frac{p}{q_1 * q_2} \geq 0$ such that the functional gradient $\frac{\delta D}{\delta q}$ is $\beta$- Lipschitz, that is $\left| \frac{\delta D}{\delta q}(q_2) - \frac{\delta D}{\delta q}(q_1) \right| \leq \beta |q_2 - q_1|$. We can verify this choice of $\beta$:
\begin{align}
    \left| \frac{\delta D}{\delta q}(q_2) - \frac{\delta D}{\delta q}(q_1) \right| 
    &= \left| \frac{-p}{q_2} - \frac{-p}{q_1} \right| \\
    &= \left| \frac{p(q_2 - q_1)}{q_2 q_1} \right| \\
    &= \frac{p}{q_2 q_1}|q_2 - q_1| \\
    & \leq \beta |q_2 - q_1| 
\end{align}

Note that the boundedness assumptions are not unrealistic in practice and can translate to a bounded parameter space for a given family of distributions. 
\begin{equation}
    \KL(p \| q_i) = \KL(p \| \sum_i^k \lambda_i f_i) = O(1/k)
\end{equation}
\section{An Alternative approach to FKL-Based Boosting: Minimizing the Remainder}\label{app:fkl_boosting_alternate}

As we seek to construct an optimal proposal through the minimization of an SNIS approximation of FKL, a trade-off arises: ``should we make the distribution easier to sample from in order to minimize the SNIS variance or should we bring it closer to the target in order to improve the worst-case IS estimation error?''
In particular, the closer the proposal gets to a multimodal target, the harder it may be to sample from. Therefore, this trade-off translates to two distinct approaches for the greedy additive construction of an optimal proposal mixture distribution.

The first approach described in Section \ref{sec:fkl_boosting} is the most straightforward as it minimizes the forward KL between the mixture $q_i$ and the target $p$ while holding the parameters of previously-learned mixture components fixed.

Alternatively, define the remainder distribution at iteration $i$ as $r_i(\theta) = \frac{p(\theta|x)}{q_i(\theta)}$. A second approach is to minimize the FKL between each new component and $r_i$, which may be simpler with fewer modes than $p$:
\begin{equation*}
    \underset{f_i}{\argmin}\;\KL(r_i\|f_i) = \underset{f_i}{\argmin}\; \KL\left(\frac{p_i}{q_{i-1}}\bigg\|f_i\right).
\end{equation*}
The mixture weight can be estimated in this scenario for each mixture component by gradient descent using the gradient with respect to FKL (see Appendix~\ref{sec:weight_grad}).

This second approach is appealing because, at each boosting iteration, $r_i$ becomes shallower with fewer modes which makes it easier to sample from than the multimodal proposal $q_i$. This approach can also be motivated by gradient boosting \citep{friedman2001greedy} or matching pursuit \citep{mallat1993matching} where one seeks to identify the mixture component that best fits the functional residual. Furthermore, this approach might be less prone to degeneracy. In fact, a derivation of this approach for the reverse KL, in Appendix~ \ref{sec:remainder_entropy}, identifies intrinsic entropy regularization which is often incorporated ad-hoc in similar objectives.

\section{Additional Simulation Experiment Results}

\begin{figure}[!h]
    \centering
    \includegraphics[width=1\linewidth]{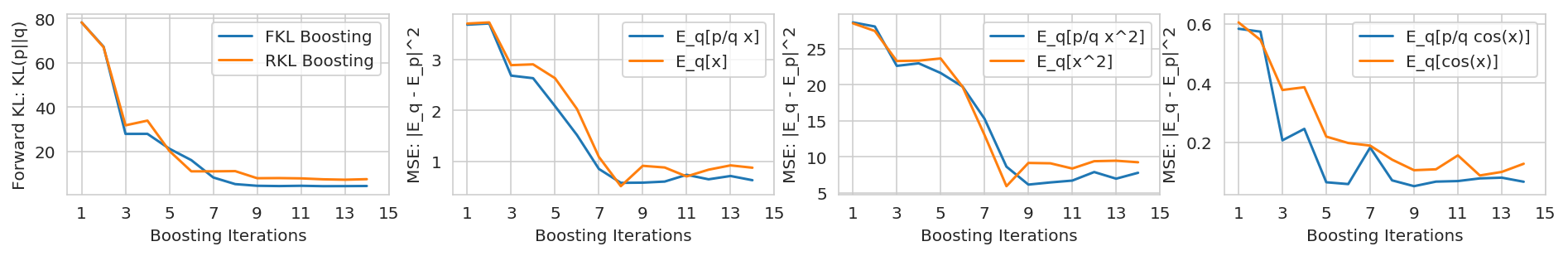}
    \caption{
    Evolution of (exact) FKL divergence and the mean-squared error of moment estimation (using samples from the FKL solution) on the task of estimating a 2-dimensional GMM of 20 components \citep{ma2019irreversible}.
    }
    \label{fig:20mog_res}
\end{figure}

Fig.~\ref{fig:20mog_res} provides moment estimation results for the simulation with well-separated modes on a mixture of 20 2-dimensional Gaussians.

\section{Additional Real Data experiment details and results}\label{app:experiments}

We provide additional details and results for the experiments on real data.

\subsection{Parameter transformations}
For the covariance matrices of each component, we optimize over the square root of diagonal matrices to ensure the non-negativity of the final diagonal covariance estimate.

To ensure non-negative or zero mixture weights, we optimize over the logits of the weights from which we recover the final weights by logistic transformation.

% \subsection{Implementation details}
% We provide additional implementation details for optimizing the variational approximation of the target distribution and sampling from the variational approximation.

% \textbf{Optimization details:} When optimizing each boosting component, we use the ADAM optimizer with a fixed learning rate and Monte Carlo gradients based on a fixed number of samples. In each boosting iteration, all mixture weights were fully re-optimized using simplex-projected gradient descent and Eq.~\eqref{eq:weight_grad}. The covariance matrices for each component were parameterized to be diagonal with positive entries.
% Our implementation leverages the automatic differentation package Autograd \citep{maclaurin2015}. 

\subsection{Hyperparameters}\label{app:hparams}
Hyperparameters for both the RKL and FKL boosting methods include the learning rate for the mean and learning rate for the covariance matrix when optimizing each boosting component. These were each tuned between $\{0.0001,0.001,0.01,0.1\}$. The number of steps for each boosting component was tuned between 200 and 1000, and the number of samples for each gradient computation was tuned between $\{25, 50, 100, 200\}$. The variance $\sigma$ when initializing the covariance matrix for each component was tuned between $\{0.0001,0.0005,0.001,0.005,0.01\}$.

%The best hyperparameter sets across methods had roughly the same number of computational budget (i.e. number of gradient steps $\times$ number of samples per gradient).

For evaluation, we draw 6000 parameter samples from the final mixture of Gaussians from which we compute the final metrics. Increasing this number to 50000 did not lead to a significant change and posed a strain on computational resources.

For the comparison to HMC, each HMC chain was initialized with a sample drawn from $\mathcal{N}(0,\sigma^2 I)$, with $\sigma = 0.01$. For all BLR tasks, the HMC comparison was run using an adaptive step size schedule with a starting step size of 1.0, 1000 burn-in steps, and 800 adaptation steps.  For the BNN tasks, the HMC comparison was run using a fixed step size, tuned between $\{0.001, 0.01\}$.

% \textbf{Sampling from the mixture:} To draw samples from the Gaussian mixture variational approximation of the target distribution, we draw a sample from component $i$ (with mean $\mu_i$ and covariance $\Sigma_i$) with probability $\lambda_i$.

\subsection{Datasets}
We use four datasets from UCI, listed in Table \ref{tab:datasets}. Table \ref{tab:datasets} reports the number of attributes in the dataset, or the dimensionality of the input $x$, but the actual dimensionality of the sampling problem in each experiment is higher than the number of attributes depending on the size of the weight vector $w$, and the additional variance parameters given in Sections \ref{sec:blr_gaussian} and \ref{sec:blr_heavy}. All tasks are regression tasks.
\begin{table}[!ht]
\centering
\resizebox{\columnwidth}{!}{
\begin{tabular}{|c|c|c|}
\hline
 Dataset name &  \# attributes & \# examples \\
\hline
\textit{wine} \citep{uci_wine} & 11 & 4898 \\
\hline
\textit{boston} \citep{uci_boston} & 13 & 506 \\
\hline
\textit{concrete} \citep{uci_concrete} & 8 & 1030 \\
\hline
% \textit{energy} \citep{uci_energy} & 8 & 768 \\
% \hline
% \textit{yacht} \citep{uci_yacht} & 6 & 308 \\
% \hline
\textit{power} \citep{uci_power} & 4 & 9568 \\
\hline
\end{tabular}
}
\caption{\label{tab:datasets}Datasets used in experiments.}
\end{table}

\subsection{Computational considerations}
In our experiments we observed that even in high dimensions, there do not seem to be significant computational differences between our method for optimizing the FKL and optimizing the RKL. The reported results use the same number of IS samples and optimization iterations for both RKL-VI and FKL-VI. In terms of wall clock time, we evaluate the highest dimensional experiment using BNNs on the Boston dataset ($d=753$). FKL VI had a wall clock time of 783.10 seconds and RKL VI had a wall clock time of 862.93 seconds after optimizing a single boosting component for 200 gradient steps when run on a single 8-core machine with an Intel Xeon CPU @ 2.20GHz.

\subsection{Additional experiment results}

In \citep{miller2017}, the posterior predictive distribution is simply estimated as an average over samples from the posterior $p(\theta | \mathcal{D}_{\text{train}})$ using variational boosting. In Tables \ref{tab:blr_uci_app}, \ref{tab:bnn_uci_app} and \ref{tab:heavy_blr_uci_app}, we report the results for the RKL boosting methods where the posterior predictive distribution is computed without importance sampling (as in Eq.~\eqref{eq:iwll}), and is instead computed by directly averaging over samples from the variational distribution, as in Eq.~\eqref{eq:vbll}.

\begin{equation}\label{eq:vbll}
p(y | x^*, \mathcal{D}_{\text{train}}) \approx \frac{1}{L} \sum_{l=1}^L p(y | x^*, \theta^{(l)}), \quad \theta^{(l)} \sim p(\theta | \mathcal{D}_{\text{train}})
\end{equation}

We also report results for HMC with 3 chains run in parallel. To compute the final predictive log probabilities, 2000 samples were drawn from each chain, and the predictive log probability was averaged over all 6000 combined samples using Eq.~\eqref{eq:vbll}.

%%%%%%%%% OLD TABLE %%%%%%%%
% \begin{table*}[!ht]
% \centering
% \begin{tabular}{|c|c|c|c|c|}
% \hline
% Method & Wine {\footnotesize (d=14)} & Boston {\footnotesize (d=16)}& Concrete {\footnotesize (d=11)} & Power {\footnotesize (d=7)} \\
% \hline
% {\footnotesize HMC (3 chains)} & -0.998	{\footnotesize ($\pm$ 0.067)}& -2.823 {\footnotesize ($\pm$ 0.084)}& -3.755 {\footnotesize ($\pm$ 0.051)}& -2.942 {\footnotesize ($\pm$ 0.039)} \\
% \hline
% {\footnotesize RKL VI (no IS)} & -0.998 {\footnotesize ($\pm$ 0.057)}& -2.804 {\footnotesize ($\pm$ 0.089)}& -3.755 {\footnotesize ($\pm$ 0.051)}& -2.956 {\footnotesize ($\pm$ 0.033)}\\
% {\footnotesize RKL VB 2 (no IS)} & -0.996 {\footnotesize ($\pm$ 0.057)}& -2.816 {\footnotesize ($\pm$ 0.086)}& -3.755 {\footnotesize ($\pm$ 0.050)}& -3.121 {\footnotesize ($\pm$ 0.019)}\\
% {\footnotesize RKL VB 3 (no IS)} & -0.995 {\footnotesize ($\pm$ 0.058)}& -2.818 {\footnotesize ($\pm$ 0.085)}& -3.755 {\footnotesize ($\pm$ 0.051)}& 	-3.131 {\footnotesize ($\pm$ 0.016)}\\
% \hline
% \end{tabular}
% \caption{\label{tab:blr_uci_app}Predictive log probabilities on test for BLR with Gaussian prior (5 train/test splits).}
% \end{table*}

%%%%%%%% NEW TABLE %%%%%%%%%%

\begin{table*}[!ht]
\centering
\begin{tabular}{|c|c|c|c|c|}
\hline
Method & Wine {\footnotesize ($d=14$)} & Boston {\footnotesize ($d=16$)}& Concrete {\footnotesize ($d=11$)} & Power {\footnotesize ($d=7$)} \\
\hline
{\footnotesize HMC (3 chains)} & -1.003	{\footnotesize ($\pm$ 0.012)}& -2.923 {\footnotesize ($\pm$ 0.035)}& -3.781 {\footnotesize ($\pm$ 0.013)}& -2.942* {\footnotesize ($\pm$ 0.017)} \\
\hline
{\footnotesize RKL VI (no IS)} & -1.003	{\footnotesize ($\pm$ 0.012)}& -2.924 {\footnotesize ($\pm$ 0.035)}& -3.780 {\footnotesize ($\pm$ 0.013)}& -2.921 {\footnotesize ($\pm$ 0.006)} \\
{\footnotesize RKL VB 2 (no IS)} & -1.003	{\footnotesize ($\pm$ 0.012)}& -2.923 {\footnotesize ($\pm$ 0.035)}& -3.781 {\footnotesize ($\pm$ 0.013)}& -2.994 {\footnotesize ($\pm$ 0.004)} \\
{\footnotesize RKL VB 3 (no IS)} & -1.003	{\footnotesize ($\pm$ 0.012)}& -2.924 {\footnotesize ($\pm$ 0.035)}& -3.781 {\footnotesize ($\pm$ 0.013)}& -2.972 {\footnotesize ($\pm$ 0.005)} \\
\hline
\end{tabular}
\caption{\label{tab:blr_uci_app}Predictive log probabilities on test for BLR with Gaussian prior (mean $\pm$ standard error over 20 train/test splits). (*Results from only 5 train/test splits due to computational constraints.) }
\end{table*}

%%%%%%%% OLD TABLE %%%%%%%%
% \begin{table*}[!h]
% \centering
% \begin{tabular}{|c|c|c|c|c|c|c|}
% \hline
% Method & Wine {\footnotesize (d=653)}& Boston {\footnotesize (d=753)} & Concrete {\footnotesize(d=503)} & Power {\footnotesize(d=303)} \\
% \hline
% {\footnotesize HMC (3 chains)} & -0.984	{\footnotesize ($\pm$ 0.077)} & -2.527 {\footnotesize ($\pm$ 0.065)} & -3.283 {\footnotesize ($\pm$ 0.105)} & -2.824 {\footnotesize ($\pm$ 0.038)} \\
% \hline
% {\footnotesize RKL VI (no IS)} & -0.997 {\footnotesize ($\pm$ 0.053)}& -2.453 {\footnotesize ($\pm$ 0.062)}& -3.389 {\footnotesize ($\pm$ 0.046)}& -2.889 {\footnotesize ($\pm$ 0.022)}\\
% {\footnotesize RKL VB 2 (no IS)}& -0.994 {\footnotesize ($\pm$ 0.054)} & -2.656 {\footnotesize ($\pm$ 0.044)}& -3.323 {\footnotesize ($\pm$ 0.048)}& -3.114 {\footnotesize ($\pm$ 0.014)}\\
% {\footnotesize RKL VB 3 (no IS)}& -0.981 {\footnotesize ($\pm$ 0.053)} & -2.685 {\footnotesize ($\pm$ 0.040)}& -3.315 {\footnotesize ($\pm$ 0.055)}& -3.163 {\footnotesize ($\pm$ 0.012)}\\
% \hline
% \end{tabular}
% \caption{\label{tab:bnn_uci_app}Predictive log probabilities on test for BNNs with Gaussian prior (5 train/test splits).}
% \end{table*}

%%%%%%% NEW TABLE %%%%%%%%
\begin{table*}[!h]
\centering
\begin{tabular}{|c|c|c|c|c|c|c|}
\hline
Method & Wine {\footnotesize ($d=653$)}& Boston {\footnotesize ($d=753$)} & Concrete {\footnotesize($d=503$)} & Power {\footnotesize($d=303$)} \\
\hline
{\footnotesize HMC (3 chains)} & -0.988	{\footnotesize ($\pm$ 0.014)}& -2.706 {\footnotesize ($\pm$ 0.093)}& -3.279 {\footnotesize ($\pm$ 0.019)}& -2.824* {\footnotesize ($\pm$ 0.017)} \\
\hline
{\footnotesize RKL VI (no IS)} & -0.991 {\footnotesize ($\pm$ 0.015)} & -2.858 {\footnotesize ($\pm$ 0.019)}& -3.230 {\footnotesize ($\pm$ 0.015)}& -2.850 {\footnotesize ($\pm$ 0.009)} \\
{\footnotesize RKL VB 2 (no IS)}& -0.990 {\footnotesize ($\pm$ 0.015)} & -2.835 {\footnotesize ($\pm$ 0.020)}& -3.231 {\footnotesize ($\pm$ 0.015)}& -2.943 {\footnotesize ($\pm$ 0.011)} \\
{\footnotesize RKL VB 3 (no IS)}& -0.983 {\footnotesize ($\pm$ 0.014)} & -2.753 {\footnotesize ($\pm$ 0.015)} & -3.232 {\footnotesize ($\pm$ 0.015)}& -2.997 {\footnotesize ($\pm$ 0.011)} \\
\hline
\end{tabular}
\caption{\label{tab:bnn_uci_app}Predictive log probabilities on test for BNNs with Gaussian prior (mean $\pm$ standard error over 20 train/test splits). (*Results from only 5 train/test splits due to computational constraints.)}
\end{table*}

%%%%%%% OLD TABLE %%%%%%%%%
% \begin{table*}[!ht]
% \centering
% \begin{tabular}{|c|c|c|c|c|}
% \hline
% Method & Wine {\footnotesize (d=13)} & Boston {\footnotesize (d=15)}& Concrete {\footnotesize (d=10)} & Power {\footnotesize (d=6)} \\
% \hline
% {\footnotesize HMC (3 chains)} & -1.009 {\footnotesize ($\pm$ 0.066)} & -2.971 {\footnotesize ($\pm$ 0.257)} & -3.787 {\footnotesize ($\pm$ 0.058)} & -2.942 {\footnotesize ($\pm$ 0.039)} \\
% \hline
% {\footnotesize RKL VI (no IS)} & -0.996 {\footnotesize ($\pm$ 0.068)}& -2.827 {\footnotesize ($\pm$ 0.084)}& -3.755 {\footnotesize ($\pm$ 0.051)}& -2.949 {\footnotesize ($\pm$ 0.032)}\\
% {\footnotesize RKL VB 2 (no IS)} & -0.994 {\footnotesize ($\pm$ 0.070)}& -2.826 {\footnotesize ($\pm$ 0.086)}& -3.754 {\footnotesize ($\pm$ 0.049)}& -3.152 {\footnotesize ($\pm$ 0.018)}\\
% {\footnotesize RKL VB 3 (no IS)} & -0.995 {\footnotesize ($\pm$ 0.069)}& -2.827 {\footnotesize ($\pm$ 0.086)}& -3.754 {\footnotesize ($\pm$ 0.049)}& -3.083 {\footnotesize ($\pm$ 0.025)}\\
% \hline
% \end{tabular}
% \caption{\label{tab:heavy_blr_uci_app}Predictive log probabilities on test for BLR with heavy tailed prior (5 train/test splits).}
% \end{table*}

%%%%%%% NEW TABLE %%%%%%
\begin{table*}[!ht]
\centering
\begin{tabular}{|c|c|c|c|c|}
\hline
Method & Wine {\footnotesize ($d=13$)} & Boston {\footnotesize ($d=15$)}& Concrete {\footnotesize ($d=10$)} & Power {\footnotesize ($d=6$)} \\
\hline
{\footnotesize HMC (3 chains)} & -1.008	{\footnotesize ($\pm$ 0.012)}& -3.085 {\footnotesize ($\pm$ 0.056)}& -3.824 {\footnotesize ($\pm$ 0.023)}& -2.942* {\footnotesize ($\pm$ 0.017)} \\
\hline
{\footnotesize RKL VI (no IS)} & -1.002	{\footnotesize ($\pm$ 0.012)}& -2.915 {\footnotesize ($\pm$ 0.034)}& -3.780 {\footnotesize ($\pm$ 0.013)}& -2.923 {\footnotesize ($\pm$ 0.006)} \\
{\footnotesize RKL VB 2 (no IS)} & -1.001	{\footnotesize ($\pm$ 0.013)}& -2.920 {\footnotesize ($\pm$ 0.034)}& -3.780 {\footnotesize ($\pm$ 0.013)}& -2.982 {\footnotesize ($\pm$ 0.003)} \\
{\footnotesize RKL VB 3 (no IS)} & -1.001	{\footnotesize ($\pm$ 0.013)}& -2.920 {\footnotesize ($\pm$ 0.034)}& -3.781 {\footnotesize ($\pm$ 0.013)}& -2.961 {\footnotesize ($\pm$ 0.005)} \\
\hline
\end{tabular}
\caption{\label{tab:heavy_blr_uci_app}Predictive log probabilities on test for BLR with heavy tailed prior (mean $\pm$ standard error over 20 train/test splits). (*Results from only 5 train/test splits due to computational constraints.)}
\end{table*}

\end{document}